\documentclass[letterpaper, 10 pt, journal, twoside]{IEEEtran}

\IEEEoverridecommandlockouts                              

\newcommand{\mbs}[1]{\ensuremath{\boldsymbol{#1}}}
\DeclareMathAlphabet{\mbf}{OT1}{ptm}{b}{n}
\newcommand{\mc}[1]{\ensuremath{\mathcal{#1}}}

\usepackage{xcolor}
\usepackage{amsmath}
\usepackage{amssymb}
\usepackage{mathtools}
\usepackage[backend=bibtex,bibstyle=ieee,citestyle=numeric-comp,doi=false,isbn=false]{biblatex}
\usepackage{xspace}
\usepackage{multirow}
\usepackage{booktabs}
\usepackage{placeins}
\usepackage[nocomma]{optidef}
\usepackage{cancel}

\newcommand{\InnerProduct}[2]{{\left \langle #1, #2\right \rangle}}
\newcommand{\rank}{{\ensuremath{\mathrm{rank}}}}
\newcommand{\diag}{{\ensuremath{\mathrm{diag}}}}

\newcommand{\trans}{{\ensuremath{\mathsf{T}}}}
\newcommand{\mbfbar}[1]{\ensuremath{\bar{\mbf{#1}}}}
\newcommand{\mbfhat}[1]{\ensuremath{\hat{\mbf{#1}}}}
\newcommand{\mbfcheck}[1]{\ensuremath{\check{\mbf{#1}}}}

\newcommand{\rnums}{\mathbb{R}}
\newcommand{\norm}[1]{\ensuremath{\left\Vert#1\right\Vert}}

\newcommand{\trace}{{\ensuremath{\mathrm{tr}}}}

\newcommand{\eye}{\mbf{1}}

\newcommand{\Gaussian}[1]{\mc{N}\left(#1\right)}

\newcommand{\ContVar}{{\mbs{\Xi}}}
\newcommand{\ContVarCol}{{\mbs{\xi}}}
\newcommand{\ContVarIdx}{{k}}
\newcommand{\ContVarIdxMax}{{K}}

\newcommand{\MatrixHomVar}{\mbf{H}}
\newcommand{\OverallRotVar}{{\mbf{C}}}
\newcommand{\OverallPosVar}{{\mbf{r}}}

\newcommand{\PositiveSemidefiniteGeq}{{\ \succcurlyeq \ }}

\newcommand{\OverallContState}{{\mc{X}}}

\newcommand{\discVar}{{\theta}}
\newcommand{\discVarAug}{{\tilde{\discVar}}}
\newcommand{\discVarAugMbs}{{\tilde{\mbs{\discVar}}}}
\newcommand{\discVarIdx}{i}
\newcommand{\NumDiscVar}{{n_{\theta}}}
\newcommand{\LandmarkIdx}{j}
\newcommand{\NumSteps}{{N}}
\newcommand{\TimeIdx}{i}

\newcommand{\NumLandmarks}{{n_\ell}}

\newcommand{\NumXCols}{{n_x}}

\newcommand{\LandmarkMeasIdx}{{k}}
\newcommand{\NumLandmarkMeas}{{K}}
\newcommand{\LndmrkMeas}{{\mbf{y}}}

\newcommand{\mbsxi}{{\mbs{\xi}}}
\newcommand{\lndmrk}{{\mbs{\ell}}}

\newcommand{\mbsTheta}{{\mbs{\Theta}}}

\newcommand{\slice}[1]{{\textrm{slice}(#1)}}
\newcommand{\sliced}[2]{{\textrm{slice}_{#2}(#1)}}
\newcommand{\textF}{{\text{F}}}

\newcommand{\sliceMat}[3]{{#1 [\slice{#2}, \slice{#3}]}}
\newcommand{\slicedMat}[4]{{#1 [\sliced{#2}{#4}, \sliced{#3}{#4}]}}
\newcommand{\stddev}{{\sigma}}

\newcommand{\NumPosesLIW}{{n_{\text{poses}}}}
\newcommand{\NumLandmarksLIW}{{n_{\text{lndmrk}}}}
\newcommand{\PoseSpacingLIW}{{\Delta t}}

\newcommand{\mosek}{\texttt{MOSEK}\xspace}

\newcounter{daggerfootnote}

\colorlet{ColorVariable}{black}
\newcommand{\revision}[1]{{\textcolor{ColorVariable}{#1}}}
\colorlet{ColorVariableForbes}{black}
\newcommand{\revisionForbes}[1]{{\textcolor{ColorVariableForbes}{#1}}}
\usepackage[nocomma]{optidef}
\usepackage{diagbox}
\usepackage{graphicx}

\title{\LARGE \bf
Globally Optimal Data-Association-Free Landmark-Based Localization Using Semidefinite Relaxations
}
\author{Vassili Korotkine, Mitchell Cohen, and James Richard Forbes$^1$
\thanks{Manuscript received: April 10, 2025; Revised July 3, 2025; Accepted July 27, 2025.}
\thanks{
This paper was recommended for publication by
Editor Sven Behnke upon evaluation of the Associate Editor and Reviewers' comments.
}
\thanks{    
This work
was supported by the Natural Sciences and Engineering Research Council
of Canada (NSERC) Alliance Grant in collaboration with Denso Corporation.}
\thanks{$^1$The authors are with the Department of Mechanical Engineering,
McGill University, Montreal, QC H3A 0C3, Canada
(e-mails: 
    \texttt{\small{vassili.korotkine@mail.mcgill.ca}},
    \texttt{\small{mitchell.cohen3@mail.mcgill.ca}}, \texttt{\small{james.richard.forbes@mcgill.ca}}).}%
  \thanks{Digital Object Identifier (DOI): see top of this page.}%
}

\begin{document}

\markboth{IEEE Robotics and Automation Letters. Preprint Version. Accepted July, 2025}
{Korotkine \MakeLowercase{\textit{et al.}}: 
Optimal Data-Association-Free Localization Using SDPs}

\maketitle

\begin{abstract}
    This paper proposes a semidefinite relaxation for landmark-based localization with unknown data associations in planar environments. 
    The proposed method simultaneously solves for the optimal robot states and data associations in a globally optimal fashion. 
    Relative position measurements to 
    \revision{a fixed set of known landmarks}
    are used, but the data association is unknown in 
    that the robot does not know which landmark each measurement is generated from. 
    The relaxation is shown to be tight in a majority of cases for moderate noise levels. 
    The proposed algorithm is compared to local Gauss-Newton baselines initialized at the dead-reckoned trajectory, 
    and is shown to significantly improve convergence to the problem's global optimum in simulation and experiment. 
    Accompanying software and supplementary material can be found at \url{https://github.com/decargroup/certifiable_uda_loc}.
\end{abstract}
\FloatBarrier
\section{Introduction and Related Work}
\label{sec:intro}
\begin{figure}
    \centering
    \includegraphics[width=\columnwidth]{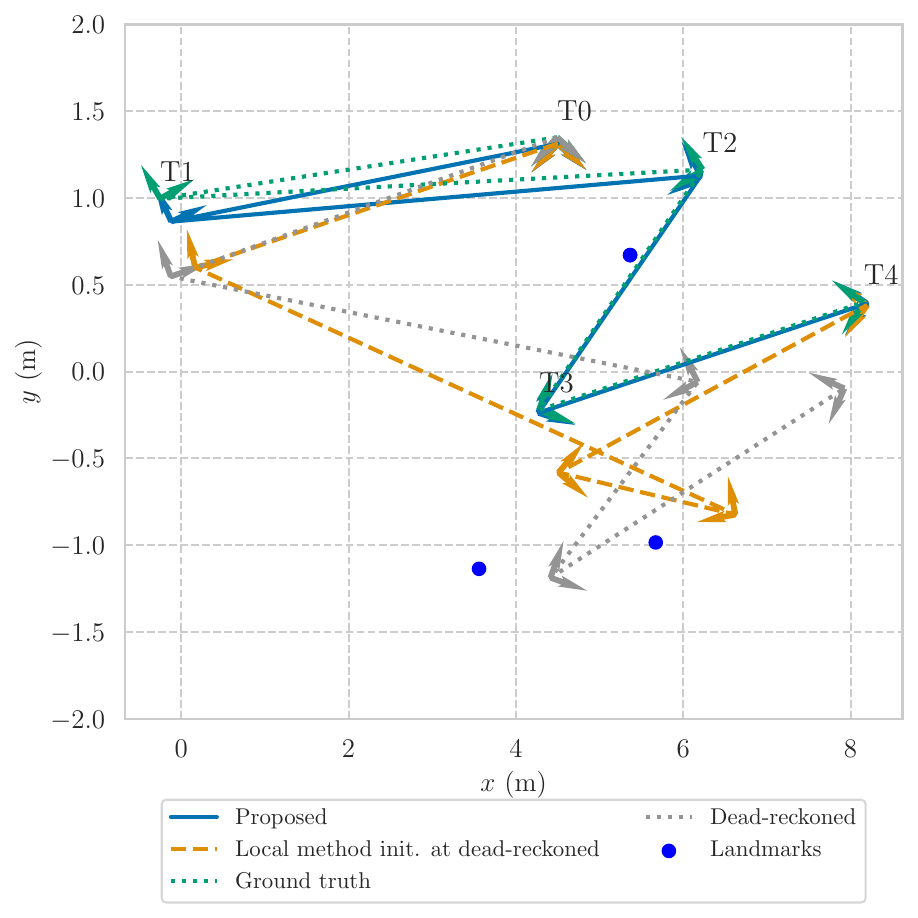}
    \vspace{-7mm}
    \caption{
        Algorithm comparison on a subsequence of the Lost in the Woods dataset. 
        The proposed SDP method converges to a solution with the correct data associations, while the local method
        initialized using dead-reckoning is unable to find the globally optimal data associations, leading to an incorrect
        robot trajectory estimate. The poses are labeled $\text{T}_i$, where $i$ corresponds to the time index. 
        Adjacent poses are 20 seconds apart.  
    }
    \label{fig:intro_result_fig}
\end{figure}
\IEEEPARstart{E}{stimating} the state of a robot from noisy and
incomplete sensor data is a central task associated with autonomy. 
In the landmark-based localization task, the robot infers its position
and orientation from measurements from landmarks with known positions. 
\revisionForbes{
State estimation methods for localization can be split into filtering methods and batch optimization methods~\cite{barfootStateEstimationRobotics2024}.
Filtering methods include Monte Carlo based localization~\cite{dellaert1999monte}, where probabilistic sampling is used to represent the posterior belief on the robot state.
} 
This work focuses on batch optimization methods, where all data over a given time window is used to estimate all the robot states
within the window.  An example is~\cite{gu2024robustjpda}, where a joint probability data association algorithm is
used to perform localization based on a high-definition map. 
\par
An important component of landmark-based localization is determining the \emph{data associations} between measurements and landmarks,
where a correspondence must be drawn between the measurements received and the landmark they are assumed to be received from.
The data associations can either be solved for before, or simultaneously with, the robot states. 
Solving for the data associations before the robot states can be done with a front-end sensor dependent system,
such as by tracking specific camera pixel coordinates between frames~\cite{shi1994features},
or it can be done in the optimization
back-end by using the provided data.
\revision{
In practice, the need to explicitly estimate data associations in the backend can arise from landmark
detections that do not provide rich information, as is common in semantic object-level SLAM~\cite{doherty2020probabilistic},
or failures due to incorrect data association in the frontend \cite{zhang2023dataassociationslam}.}
Algorithms such as 
joint compatibility branch and bound~\cite{neira2001jcbb}, combined constraint data association~\cite{bailey2002mobile},
and the CLEAR alignment and rectification algorithm~\cite{fathian2020clustering} fall into this category.
In all of the above algorithms~\cite{neira2001jcbb,bailey2002mobile,fathian2020clustering}, the data associations are solved for \emph{before} solving for the robot states with the data associations fixed,
which can result in a suboptimal solution if the initial data associations are incorrect. 
%
\par
Another approach consists of solving for the data associations \emph{simultaneously} with the robot states, a problem that is
NP-hard and combinatorial in nature~\cite{zhang2023dataassociationslam}. 
When applied to this problem formulation, the Max-Mixture approach of~\cite{olsonInferenceNetworksMixtures2013}
iterates through optimization steps selecting the most likely data association at each step. 
The data-association-free landmark-based simultaneous localization and mapping (SLAM) algorithm of~\cite{zhang2023dataassociationslam}
simultaneously solves for robot states, landmark positions, and the data associations in an inner loop using the Max-Mixture algorithm, while solving
for the number of landmarks in an outer solver loop. 
Probabilistic data association
in a Max-Mixture framework is also presented in~\cite{doherty2020probabilistic} for the application of semantic SLAM.
DC-SAM~\cite{doherty2022dcsam} is a general purpose method for problems involving
continuous and discrete variables that involves alternation of solving for the continuous variables
and then solving for the discrete variables. 
\par
The algorithms 
 described above that simultaneously solve for the data associations and robot state
 are inherently \emph{local} in nature, as they require an initial guess that is refined to
obtain the optimal solution. The quality of the solution is impacted by the initial guess. Furthermore,
global optimality of the algorithm is not guaranteed. The incorrect minimum may be reached without any indication that it is the local minimum. 
\par
The present work focuses on globally optimal solutions obtained from semidefinite relaxations of the problem. 
Globally optimal methods based on semidefinite relaxations are applicable to problems that may be formulated 
as polynomial optimization problems, which are written as quadratically constrained
quadratic programs (QCQPs) using a suitable change of variables. The QCQP
admits a semidefinite program (SDP) relaxation, termed Shor's relaxation~\cite[Chap.~3]{yang2024semidefinite}.
The SDP relaxation may be solved to global optimality, and if the result is feasible for the non-relaxed problem,
it is guaranteed to be the global optimum of the non-relaxed problem.
SDP relaxations have been used to treat outlier rejection~\cite{lajoie2019perceptual, yang2023outlier}, 
where the algorithm must decide if a given measurement is an outlier or not, a choice that is encoded using discrete boolean states that
are solved for alongside the continuous robot states. 
%
%
%
%
\emph{The contribution of this paper may thus be stated as a relaxation-based globally optimal method, in an optimization cost sense, for the
data-association-free planar \revisionForbes{landmark-based} localization problem \revision{where the number of landmarks is known}.}
To the best of the authors' knowledge, this is the first such method that scales in polynomial time.
An example result is presented in Fig.~\ref{fig:intro_result_fig}, where for an example with five poses
and three landmarks, the proposed method obtains the globally optimal solution with the correct data associations,
unlike the local method initialized using dead-reckoning. 
\revision{The method is applicable to ground vehicle localization, in situations where relative pose measurements as well as measurements to 
known landmarks are available.}
The
implementation is open-sourced at \url{https://anonymous.4open.science/r/certifiable_uda_loc-9BCF/README.md}.
\revision{Supplementary material at the same URL contains further details on the cost and constraint matrix setup. }
\par
The rest of this paper is organized as follows. Sec.~\ref{sec:problem_setup}
describes the problem setup of landmark-based localization with unknown data associations.
Sec.~\ref{sec:proposed_approach} describes the proposed SDP relaxation.
Sec.~\ref{sec:results} presents results on simulated and experimental data, while
Sec.~\ref{sec:conclusion} concludes and provides directions for future work.
%
%
\begin{table}[htbp]\caption{Notation Table}
    \begin{center}
    \begin{tabular}{r c p{6cm} }
    \toprule
    \multicolumn{3}{c}{\underline{Indices}}\\
    $\TimeIdx$ & $\triangleq$ & Timestep, sometimes generic index\\
    $\NumSteps$ & $\triangleq$ & Number of timesteps\\
    $\LandmarkIdx$ & $\triangleq$ & Landmark index, sometimes generic index \\
    $\NumLandmarks$ & $\triangleq$ & Number of landmarks \\
    $\LandmarkMeasIdx$ & $\triangleq$ & Index of measurement received at given timestep, sometimes generic index \\
    $\NumLandmarkMeas$ & $\triangleq$ & Number measurements received at given timestep \\
    $\NumDiscVar$ & $\triangleq$ & Number of discrete problem variables \\
    $n_x$ & $\triangleq$ & Number of columns in optimization variable $\mbf{X}$\\
    \multicolumn{3}{c}{}\\
    \multicolumn{3}{c}{\underline{Quantities}}\\
    \revision{$\eye$} & \revision{$\triangleq$} & \revision{Identity matrix with unspecified dimension} \\ 
    \revision{$\eye^{n \times n}$} & \revision{$\triangleq$} & \revision{Identity matrix of dimension $n \times n$}  \\ 
    $\mbf{T}_\TimeIdx \in SE(3)$ & $\triangleq$ & Robot pose at time index $\TimeIdx$ \\
    $\mc{X} \in {SE(3)}^\NumSteps$ & $\triangleq$ & Tuple of all robot poses \\
    $\discVar \in \{0, 1\}$ & $\triangleq$ & Single discrete problem variable \\
    $\mbs{\discVar} \in \rnums^\NumDiscVar$ & $\triangleq$ & All discrete problem variables stacked into column matrix \\
    $\lndmrk_\LandmarkIdx  \in \rnums^2$ & $\triangleq$ & $\LandmarkIdx$'th landmark position \\
    $\mbf{H} \in \rnums^{2\times 2}$  & $\triangleq$ &  Homogenization variable \\
    $\mbf{X}\in \rnums^{2\times \NumXCols}$ & $\triangleq$ & Matrix variable with variables of interest concatenated  \\
    $\ContVar \in \rnums^{2\times (2+3\NumSteps)}$ & $\triangleq$ & Concatenation of continuous problem variables  \\
    $\mbf{Z} \in \rnums^{\NumXCols \times \NumXCols}$ & $\triangleq$ & Optimization matrix variable in Shor's relaxation \\
    \multicolumn{3}{c}{}\\
    \multicolumn{3}{c}{\underline{Operators}}\\
    $\otimes$ & $\triangleq$ & Kronecker product \\
    $\InnerProduct{\mbf{A}}{\mbf{B}}$ & $\triangleq$ & Matrix inner product. For $\mbf{A}$, $\mbf{B}\in \rnums^{m\times n}$, 
    $\InnerProduct{\mbf{A}}{\mbf{B}}=\trace(\mbf{A}\mbf{B}^\trans)$ \\
    \revision{$\mc{N}(\mbf{x}; \mbfbar{x}, \mbf{P})$} & \revision{$\triangleq$} & \revision{A multivariate Gaussian distribution in $\mbf{x}$, with mean $\mbfbar{x}$ and covariance $\mbf{{P}}$} \\
    \bottomrule
    \end{tabular}
    \end{center}
    \label{tab:notation_table}
    \end{table}
\section{Problem Setup}
\label{sec:problem_setup}
For convenience, commonly used notation is summarized in Tab.~\ref{tab:notation_table}.
The state of interest $\OverallContState$ is given by 
$\mc{X}=\left\{\mbf{T}_1, \dots,  \mbf{T}_{\TimeIdx}, \dots, \mbf{T}_{\NumSteps}\right\}\in  SE(2)^\NumSteps$
where the timestep index is denoted $\TimeIdx$ and $\NumSteps$ is the number of timesteps.   
The robot poses are denoted $\mbf{T}_{\TimeIdx}=(\mbf{C}_{\TimeIdx}, \mbf{r}_{\TimeIdx}) \in SE(2)$.
$\mbf{C}_{\TimeIdx} \in SO(2)$ denotes the direction cosine matrix at timestep $\TimeIdx$ describing the change of basis
from the robot frame to the inertial frame, and $\mbf{r}_{\TimeIdx}$
denotes the robot position at timestep $\TimeIdx$ resolved in the inertial frame.
At the $\TimeIdx$'th timestep, the robot receives $\NumLandmarkMeas_{\TimeIdx}$
relative landmark position measurements $\LndmrkMeas_{\LandmarkMeasIdx_{\TimeIdx}}$,
to landmarks with known positions,
$\lndmrk_\LandmarkIdx$ for $j=1, \dots, \NumLandmarks$.
The number of measurements received at the $\TimeIdx$'th timestep $\NumLandmarkMeas_{\TimeIdx}$ is typically
less than the number of landmarks $\NumLandmarks$.
The generative isotropic noise model for the relative landmark position measurements is given by
\begin{align}
    \LndmrkMeas_{\LandmarkMeasIdx_{\TimeIdx}} &= \mbf{C}_\TimeIdx^\trans  (\lndmrk_\LandmarkIdx - \mbf{r}_\TimeIdx) + \mbf{v}, \quad \mbf{v} \sim
    \Gaussian{\mbf{v}; \mbf{0}, \eye \stddev_{\text{lndmrk}}^2}, 
    \label{eq:landmark_meas_generative}
\end{align}
\revision{where the covariance $\eye \stddev_{\text{lndmrk}}^2$ is a multiple of identity scaled by variance $\eye \stddev_{\text{lndmrk}}^2$, since isotropic noise is considered.}
At timestep $i$, the data association variables are encoded as 
$\theta_{\TimeIdx \LandmarkMeasIdx_{\TimeIdx} \LandmarkIdx}\in \{0, 1\}$. 
A value of $\theta_{\TimeIdx \LandmarkMeasIdx_{\TimeIdx} \LandmarkIdx}=1$ encodes that, at timestep $\TimeIdx$, the 
$\LandmarkMeasIdx_{\TimeIdx}$'th measurement received at that timestep corresponds to landmark $\lndmrk_\LandmarkIdx$.
The data association variables are further constrained by 
\begin{align}
    \sum_{\LandmarkIdx=1}^{\NumLandmarks} \discVar_{\TimeIdx \LandmarkMeasIdx \LandmarkIdx} =1, \quad \forall {\TimeIdx, \LandmarkMeasIdx},
    \label{eq:disc_var_sum_constraint}    
\end{align}
meaning that every measurement comes from a single landmark. 
All the $\NumDiscVar = \sum_{\TimeIdx=1}^\NumSteps \sum_{\LandmarkMeasIdx_{\TimeIdx}=1}^{\NumLandmarkMeas_{\TimeIdx}} \NumLandmarks$
data association variables may be stacked into a column vector $\mbs{\discVar}$.
The problem formulation may thus be stated as
\begin{mini*}|l|
    {\mc{X}, \mbs{\discVar}}{
     J_\text{odom}(\mc{X}) + J_\text{prior}(\mc{X}) + J_{\text{uda, lndmrk}}(\mc{X}, \mbs{\discVar}), 
    }{}{}\tag{\text{Localization}}
    \label{prob:original_problem}
\end{mini*}
with $\mc{X}\in SE(2)^\NumSteps$ and $\mbs{\discVar} \in \rnums^{\NumDiscVar}$.
The first term is the odometric cost~\cite{holmes2023efficient} given by
\begin{align}
    J_\text{odom}(\mc{X})&= 
    \sum_{\TimeIdx}^{\NumSteps}
    \kappa_\text{odom}
    \norm{\mbf{C}_\TimeIdx-\mbf{C}_\TimeIdx \Delta \mbf{C}_\TimeIdx }_{\textF}^2 + 
    \nonumber
    \\
    & \quad 
    \frac{1}{\stddev_{r, \text{odom}}^2}
    \norm{\mbf{r}_{\TimeIdx+1} - \mbf{r}_\TimeIdx-\mbf{C}_{k} \Delta \mbf{r}_i}_2^2,
    \label{eq:loss_odom}
\end{align}
where $\Delta \mbf{C}_i$ is the relative direction cosine measurement and $\Delta \mbf{r}_i$ is the relative position measurement. 
The prior cost inserts a prior on the first pose and is given by
\begin{align}
    J_\text{prior}(\mc{X})&= 
    \kappa_{\text{prior}}
    \norm{\mbf{C}_1-\mbfcheck{C}_1}_{\textF}^2 + 
    \frac{1}{\stddev_{r, \text{prior}}^2}
    \norm{\mbf{r}_1-\mbfcheck{r}_1}_2^2,
    \label{eq:loss_prior}
\end{align}
where $\mbfcheck{C}_1$ and $\mbfcheck{r}_1$ are the prior orientation and position, respectively. 
The log-likelihood residual arising from~\eqref{eq:landmark_meas_generative} can be written as~\cite{holmes2023efficient}, 
\begin{align}
    r(\mbf{C}_\TimeIdx, \mbf{r}_\TimeIdx; \lndmrk_\LandmarkIdx, \LndmrkMeas)
     & =
     \frac{1}{\stddev_{\text{lndmrk}}^2}
    \norm{
    (\lndmrk_\LandmarkIdx - \mbf{r}_\TimeIdx) - \mbf{C}_\TimeIdx \LndmrkMeas
    }_2^2,
    \label{eq:single_known_landmark_loss}
\end{align}
Such that the unknown data association landmark measurement cost $J_{\text{uda, lndmrk}}(\mc{X}, \mbs{\discVar})$ becomes
\begin{align}
    J_{\text{uda, lndmrk}}(\mc{X}, \mbs{\discVar}) &= 
    \sum_{\TimeIdx=1}^{\NumSteps}
    \sum_{\LandmarkMeasIdx_{\TimeIdx}=1}^{\NumLandmarkMeas_{\TimeIdx}}
    \sum_{\LandmarkIdx=1}^{\NumLandmarks}
    \discVar_{\TimeIdx \LandmarkMeasIdx \LandmarkIdx}
    r(\mbf{C}_\TimeIdx, \mbf{r}_\TimeIdx; \lndmrk_\LandmarkIdx, \LndmrkMeas_{\LandmarkMeasIdx_{\TimeIdx}}). 
    \label{eq:uda_landmark_measurement_cost}
\end{align}
The outer loop is over the time indices $\TimeIdx$.
The middle loop is over the $\LandmarkMeasIdx$'th received measurement at timestep $\TimeIdx$, $\LndmrkMeas_{\LandmarkMeasIdx_{\TimeIdx}}$. 
The inner loop is over the possible data associations. 
The SDP formulation used for treating unknown data associations is very similar to the certifiable outlier rejection case~~\cite{lajoie2019perceptual, yang2023outlier}, 
However, the present work uses the assumption that a measurement came from one of
a set of landmarks, as opposed to~\cite{lajoie2019perceptual, yang2023outlier} where the measurements are assumed to follow a given
probability distribution. The outlier problem solutions~\cite{lajoie2019perceptual, yang2023outlier} add cost terms to the optimization,
while the data-association-free assumption of the present work leads to additional SDP constraints arising from the sum constraint~\eqref{eq:disc_var_sum_constraint}.
Outlier rejection using SDPs has been considered by encoding the outlier status of
measurements as boolean variables that are solved for alongside the continuous states of interest~\cite{lajoie2019perceptual, yang2023outlier}.
Cost terms that are dependent on the boolean variables that depend on the assumed outlier distribution are included in the optimization.
\revision{
Furthermore, the number of landmarks is assumed known in this problem setup. 
Thus, the problem setup essentially corresponds to the inner loop of the local algorithm proposed in \cite{zhang2023dataassociationslam}. 
A nuance of the proposed approach is in the assumption that a measurement comes from one of a set of landmarks. This assumption does not prevent many
measurements being assigned to a single landmark in a ``many-to-one'' fashion, leaving open the possibility of using
a null hypothesis landmark in future work, similarly to~\cite{poschmann2020factor}. 
}
\section{Proposed Approach}
\label{sec:proposed_approach}
The proposed approach consists of a semidefinite relaxation in the spirit of the outlier rejection approach in~\cite{lajoie2019perceptual}. 
\subsection{Mathematical Formulation}
A homogenization variable $\mbf{H}\in \{-\eye_{2\times 2}, \eye_{2\times 2}\}$, which can be enforced quadratically,
is introduced to be able to write cost and constraint terms
that are linear in $\mc{X}$ and $\mbs{\discVar}$. 
The impact of homogenization is discussed in~\cite{cifuentes2022local}. 
To fix ideas, especially related to cost functions and constraints, $\mbf{H}=\eye_{2\times 2}$
can be assumed. 
The direction cosine matrix variables may be concatenated 
to yield $\OverallRotVar=
\begin{bmatrix}
    \mbf{C}_1 & \cdots & \mbf{C}_{\TimeIdx} & \cdots & \mbf{C}_{\NumSteps} 
\end{bmatrix}$, 
while the position variables may be concatenated as 
$\OverallPosVar=\begin{bmatrix} \mbf{r}_1 & \cdots & \mbf{r}_{\TimeIdx} & \cdots & \mbf{r}_{\NumSteps}\end{bmatrix}$.
The continuous problem variable may be written
\begin{align}
    \ContVar & =
    \begin{bmatrix}
        \MatrixHomVar & \OverallRotVar & \OverallPosVar
    \end{bmatrix} \in \rnums^{2\times (2+3\NumSteps)}.
    \label{eq:cont_var_definition}
\end{align}
The columns of $\ContVar$ are denoted as $\ContVarCol_\ContVarIdx$ such that 
$\ContVar=\begin{bmatrix}
    \ContVarCol_1 & \dots & \ContVarCol_\ContVarIdx & \dots & \ContVarCol_\ContVarIdxMax  
\end{bmatrix}$. 
The odometric cost~\eqref{eq:loss_odom} may be written as $\InnerProduct{\ContVar^\trans \ContVar}{\mbf{Q}_{\text{odom}}}$
where
the matrix inner product $\InnerProduct{\cdot}{\cdot}$ is defined 
for matrices $\mbf{A}, \mbf{B}\in \rnums^{m\times n}$
as $\InnerProduct{\mbf{A}}{\mbf{B}}=\trace(\mbf{A}\mbf{B}^\trans)=\sum_{i=1}^M \sum_{j=1}^N a_{ij}b_{ij}$, and 
$\mbf{Q}_{\text{odom}}$ is a data matrix that depends on the measurements $\Delta \mbf{C}_i, \Delta \mbf{r}_i$.
Similarly, the prior cost~\eqref{eq:loss_prior} can be written as $\InnerProduct{\ContVar^\trans \ContVar}{\mbf{Q}_{\text{prior}}}$
with $\mbf{Q}_{\text{prior}}$ a data matrix that depends on the prior values. 
The \emph{known} data association loss term corresponding to a given timestep and landmark measurement~\eqref{eq:single_known_landmark_loss} can also be written
as $\InnerProduct{\ContVar^\trans \ContVar}{\mbf{Q}_{\text{lndmrk}, \TimeIdx, \LandmarkMeasIdx, \LandmarkIdx}}$
where $\mbf{Q}_{\text{lndmrk}, \TimeIdx, \LandmarkMeasIdx, \LandmarkIdx}$ depends on the $\LandmarkIdx$'th landmark position $\lndmrk_\LandmarkIdx$
as well as on the measurement received, $\mbf{y}_{\LandmarkMeasIdx_{\TimeIdx}}$.
\par
To write the SDP formulation for the unknown data association, let the optimization variable $\mbf{X}$ be defined as
\begin{align}
    \mbf{X} & =
    \begin{bmatrix}
        \MatrixHomVar & \mbs{\theta}^\trans \otimes \eye^{2\times 2} & \mbs{\theta}^\trans \otimes \mbs{\Xi} & \mbs{\Xi}
    \end{bmatrix},
    \label{eq:optimization_variable_definition}
\end{align}
where $\mbf{X}$ has $\NumXCols=(2+2\NumDiscVar+3\NumSteps(\NumDiscVar+1))$ columns such that
$\mbf{X}\in \rnums^{2\times \NumXCols}$.
The unknown data association loss~\eqref{eq:uda_landmark_measurement_cost}
may then be written as $\InnerProduct{\mbf{X}^\trans \mbf{X}}{\mbf{Q}_{\text{uda}}}$, 
which contains the individual known data association cost matrices $\mbf{Q}_{\text{lndmrk}, \TimeIdx, \LandmarkMeasIdx, \LandmarkIdx}$
to achieve equality between $\InnerProduct{\mbf{X}^\trans \mbf{X}}{\mbf{Q}_{\text{uda}}}$ and~\eqref{eq:uda_landmark_measurement_cost}.
The overall problem cost matrix may be written $\mbf{Q}=\mbf{Q}_{\text{uda}}+\mbf{Q}_{\text{odom}}+\mbf{Q}_{\text{prior}}$, and
Problem~\eqref{prob:original_problem} may be formulated as
\begin{mini*}|l|
    {\mbf{X}}{
        \sum_{i=1}^{\NumDiscVar}
    \InnerProduct{\mbf{X}^\trans \mbf{X}}{\mbf{Q}}
    }{}{}\tag{\text{Reform. Prob}}
    \label{prob:prob_in_terms_of_x}
    \addConstraint{\mbf{X}\in \{-\eye, \eye\} \times  \{\mbf{0}, \eye\}^{\NumDiscVar} \times }
    \addConstraint{}{\quad \left(SO(2)^\NumSteps \times (\rnums^2)^\NumSteps\right)^{(1+\NumDiscVar)}.}
\end{mini*}
The convention of~\cite{lajoie2019perceptual} 
for overloading the $\times$ operator
is used for defining blockwise matrix membership constraints.  
Relaxing the $\mbf{C}_i\in SO(2)$ orthonormality constraint to the orthogonal constraint $\mbf{C}_i\in O(2)$ yields 
the constraint $\mbf{X}\in \{-\eye, \eye\} \times  \{\mbf{0}, \eye\}^{\NumDiscVar} \times  \left(O(2)^\NumSteps \times (\rnums^2)^\NumSteps\right)^{(1+\NumDiscVar)}$, 
which can be written fully in terms of 
a homogenization constraint $\InnerProduct{\mbf{A}_{\text{hom}}}{\mbf{X}^\trans \mbf{X}}=1$
as well as quadratic constraints of the form $\InnerProduct{\mbf{A}_{i, \text{init}}}{\mbf{X}^\trans \mbf{X}}=0, \quad i=1, \dots, n_\text{constr, init}$.
These constraints, with constraint matrices $\mbf{A}_{i, \text{init}}$, are termed the initial constraints, as additional redundant constraints have to then
be added to tighten the relaxation. 
The orthogonality relaxation can thus be written
\begin{mini*}|l|
    {\mbf{X}}{
    \InnerProduct{\mbf{X}^\trans \mbf{X}}{\mbf{Q}}
    }{}{}\tag{\text{Orth. Relax.}}
    \label{prob:relaxation_x_1}
    \addConstraint{\InnerProduct{\mbf{A}_{\text{hom}}}{\mbf{X}^\trans \mbf{X}}}{=1}
    \addConstraint{\InnerProduct{\mbf{A}_{i, \text{init}}}{\mbf{X}^\trans \mbf{X}}}{=0, \quad i=1, \dots, n_\text{constr, init}}.
\end{mini*}
Introducing a matrix variable $\mbf{Z}$ that corresponds to $\mbf{X}^\trans \mbf{X}$ an equivalent problem to Problem~\eqref{prob:relaxation_x_1}
as
\begin{mini*}|l|
    {\mbf{Z}}{
    \InnerProduct{\mbf{Z}}{\mbf{Q}}
    }{}{}\tag{\text{Orth. Relax. Z.}}
    \label{prob:orth_relax_z}
    \addConstraint{\InnerProduct{\mbf{A}_{\text{hom}}}{\mbf{Z}}}{=1}
    \addConstraint{\InnerProduct{\mbf{A}_{i, \text{init}}}{\mbf{Z}}}{=0, \quad i=1, \dots, n_\text{constr, init}}
    \addConstraint{\rank(\mbf{Z})}{=2}. 
\end{mini*}
Relaxing the 
$\rank(\mbf{Z})=2$ constraint yields the semidefinite program, 
\begin{mini*}
    {\mbf{Z}}{\InnerProduct{\mbf{Q}}{\mbf{Z}}}{}{}\tag{\text{Rank Relax.}}
    \label{prob:relaxation}
    \addConstraint{\InnerProduct{\mbf{A}_{\text{hom}}}{\mbf{Z}}}{=1}
    \addConstraint{\InnerProduct{\mbf{A}_{i, \text{init}}}{\mbf{Z}}}{=0, \quad i=1, \dots, n_\text{constr, init}}
    \addConstraint{\mbf{Z}}{\PositiveSemidefiniteGeq 0}.
\end{mini*}
\eqref{prob:relaxation} is convex and admits a globally optimal solution in polynomial time.
\revisionForbes{
If the solution to~\eqref{prob:relaxation} has rank two, then it is feasible for~\eqref{prob:orth_relax_z}, and is thus the global optimum for~\eqref{prob:orth_relax_z}.
The relaxation is then considered \emph{tight}, specifically in the rank sense as the solution
to~\eqref{prob:relaxation} would satisfy the rank two constraint of~\eqref{prob:orth_relax_z}.
The variable $\mbf{X}$ may then be extracted from $\mbf{Z}$ and is the global optimum for~\eqref{prob:relaxation_x_1}.
In turn, if this $\mbf{X}$ is also feasible for~\eqref{prob:prob_in_terms_of_x}, such that the direction cosine matrix blocks of $\mbf{X}$ have determinant equal to one,
the globally optimal solution to~\eqref{prob:prob_in_terms_of_x}, and thus the localization problem~\eqref{prob:original_problem}, is obtained. 
}
However, tightness
is not guaranteed since the explicit rank constraint is dropped from~\eqref{prob:orth_relax_z}. 
To encourage tightness, redundant constraints of the form $\InnerProduct{\mbf{A}_{\text{red}, i}}{\mbf{Z}}=0$ are added to the relaxation to yield 
\begin{mini*}
    {\mbf{Z}}{\InnerProduct{\mbf{Q}}{\mbf{Z}},}{}{}\tag{\text{Tightened Relax.}}
    \label{prob:relaxation_red_constraints_tightened}
    \addConstraint{\InnerProduct{\mbf{A}_{\text{hom}}}{\mbf{Z}}}{=1}
    \addConstraint{\InnerProduct{\mbf{A}_{i, \text{init}}}{\mbf{Z}}}{=0, \quad i=1, \dots, n_\text{constr, init}}
    \addConstraint{\InnerProduct{\mbf{A}_{j, \text{red}}}{\mbf{Z}}}{=0, \quad j=1, \dots, n_\text{constr, red}}
    \addConstraint{\mbf{Z}}{\PositiveSemidefiniteGeq 0}.
\end{mini*}
\subsection{Redundant Constraints}
Finding the redundant constraints is problem-specific and can be difficult. A general purpose method is proposed in~\cite{duembgen2024automatically}
that exploits the fact that any constraint matrix $\mbf{A}$ lies in the nullspace, in a matrix inner product sense, of all feasible points
for the SDP. By generating many such feasible points and examining the nullspace of the resulting data matrix, redundant constraints may be obtained. 
The work of~\cite{duembgen2024automatically} proposes both a method for finding constraints for a given problem instance, \texttt{AutoTight},
as well as a method of generalizing them, \texttt{AutoTemplate}. The present work used \texttt{AutoTight} on small problem instances, which yielded interpretable
constraints that were then applied to larger problem instances. These analytical constraints are presented in the hope that
they illuminate the problem structure and can be useful to potential analytical future work. 
\par
Herein, 
given a matrix $\mbf{A}=\begin{bmatrix}
    \mbf{a}_1 & \dots & \mbf{a}_i & \dots & \mbf{a}_N
\end{bmatrix}$, the integer $i=\sliced{\mbf{a}_i}{\mbf{A}}$ is used to denote the column index corresponding to $\mbf{a}_i$ in $\mbf{A}$. 
For instance, denoting the first column of the homogenization variable $\mbf{H}$ as $\mbf{h}_1$, then
$\slicedMat{\mbf{Z}}{\mbf{h}_1}{\mbf{h}_1}{\mbf{X}}$ is used to refer to the entry of $\mbf{Z}$
that has both the row and column index equal to the position of $\mbf{h}_1$ in the optimization variable $\mbf{X}$ in~\eqref{eq:optimization_variable_definition}.
\subsubsection{Discrete Variable Constraints}
Let $\discVarAugMbs^\trans=\begin{bmatrix}
    1 & \mbs{\discVar}^\trans
\end{bmatrix}$.
These constraints arise from algebraic observations of the form
    $\discVarAugMbs^\trans
    \mbf{A}_{\theta}
    \discVarAugMbs
    = 0$
\subsubsection{Continuous Variable Constraints}
The continuous variable constraints are of the form
\begin{align}
    \InnerProduct{\mbf{A}_{\Xi}}{
        \begin{bmatrix}
            \eye \\ \ContVar^\trans
        \end{bmatrix}
        \begin{bmatrix}
            \eye & \ContVar
        \end{bmatrix}
    \label{eq:cont_var_constraints}
    }                          & =0.
\end{align}
For the robot navigation problem considered, these constraints correspond to the
direction cosine matrix parts being constrained as orthogonal. 
\subsubsection{Combining Discrete and Continuous Variable Constraints}
Constraints acting on rows and columns corresponding to combinations of discrete and continuous variables
can be formed as follows. 
First, considering
two columns of $\ContVar$, denoted $\ContVarCol_k, \ContVarCol_\ell$, as well
as two discrete variables in the problem $\discVar_i, \discVar_j$, 
a set of redundant constraints comes from the algebraic observation given by 
\begin{align}
    \ContVarCol_k^\trans \ContVarCol_\ell
    \discVarAugMbs^\trans
    \mbf{A}_{\theta}
    \discVarAugMbs
        & =
    \sum_{i,j} a_{ij}\discVarAug_i \discVarAug_j
    \ContVarCol_k^\trans \ContVarCol_\ell
    \\
        & =
    \sum_{i,j} a_{ij}
    \left(\discVarAug_i \ContVarCol_k^\trans\right)
    \left(\discVarAug_j\ContVarCol_\ell\right),
\end{align}
which corresponds to constraint matrix $\mbf{A}$ with elements
\begin{align}
    \mbf{A}[\sliced{\discVarAug_i \ContVarCol_k}{\mbf{X}}, \sliced{\discVarAug_j\ContVarCol_\ell}{\mbf{X}}] & =
    \mbf{A}_\discVarAug[
        \sliced{\discVarAug_i}{\mbs{\discVarAugMbs^\trans}}, \sliced{\discVarAug_j}{\discVarAugMbs^\trans}
    ].
\end{align}
Second, given a constraint of the form
\begin{align}
    g(\ContVar) & =
    \InnerProduct{\mbf{A}_{\ContVar}}{
        \begin{bmatrix}
            \MatrixHomVar & \ContVar
        \end{bmatrix}^\trans
        \begin{bmatrix}
        \MatrixHomVar \\ \ContVar
        \end{bmatrix}
    }                      
            = \mbf{0},
\end{align}
it can be premultiplied by $\discVar_i$ to yield
\begin{align}
    \discVar_i g(\ContVar) 
                        & =
    \InnerProduct{\mbf{A}_{\ContVar}}{
        \begin{bmatrix}
            \mbsTheta_i            & \theta_i \ContVar                            \\
            \theta_i \ContVar^\trans & (\theta_i \ContVar)^\trans (\theta_i \ContVar)
        \end{bmatrix}
    },
    \label{eq:where_bool_constraint_used_2}
\end{align}
where the boolean constraint $\theta_i^2 =\theta_i$ was used.
\revision{
In addition, moment constraints as well as constraints due to the particular 
matrix structure of~\eqref{eq:optimization_variable_definition} have to be added to
the relaxation. These constraints are described in the supplementary material. 
}
\subsection{Sparsity}
A sparse basis is used to reduce the computational footprint of the algorithm.
The $\mbs{\theta}^\trans \otimes \mbs{\Xi}$ portion of the optimization variable definition~\eqref{eq:optimization_variable_definition}
is reduced solely to the terms necessary to express the unknown data association cost~\eqref{eq:uda_landmark_measurement_cost}.
Furthermore, 
redundant 
constraints that involve subsets of the discrete variables $\discVar_i$ and continuous variable columns $\ContVarCol_k$ are reduced
to those where $\discVar_\discVarIdx$ and $\ContVarCol_\ContVarIdx$ correspond to the same timestep $\TimeIdx$. 
This is found necessary to keep the problem size and computational footprint tractable, while still yielding an
acceptably tight relaxation. 
\subsection{Extraction of Robot States From Optimization Variable}
The extraction of the robot states from the solution to~\eqref{prob:relaxation_red_constraints_tightened}
is done in two steps. 
First, the optimization variable $\mbf{X}$ of~\eqref{prob:prob_in_terms_of_x} is extracted
from $\mbf{Z}$. 
This is done by recognizing that $\mbf{Z}=\mbf{X}^\trans \mbf{X}$, and thus that the first two rows of
$\mbf{Z}$ are $\mbf{Z}[:2, :]=\mbf{H}^\trans \mbf{X}$,
and thus directly correspond to $\mbf{X}$. 
The robot states then correspond to the
continuous variable $\ContVar$, extracted from the last block matrix of $\mbf{X}$ in~\eqref{eq:optimization_variable_definition}.
The data association variables are then extracted from the $\mbs{\theta}^\trans \otimes \eye^{2\times 2}$ block of $\mbf{X}$
in~\eqref{eq:optimization_variable_definition}.
\section{Simulation and Experiments}
\label{sec:results}
The experiments aim to demonstrate that the 
global minimum of the objective in
Problem~\eqref{prob:original_problem} is attained by the proposed approach, while the local method
baseline gets trapped in local minima with incorrect data associations.
The proposed algorithm is evaluated on simulated examples as well as real-world data. 
The parameters of interest are given by relative pose measurement noise levels,
the relative landmark position measurement noise, the prior on the first pose, as well
as the number of poses and landmarks in the setup. 
The proposed method consists of solving the SDP~\eqref{prob:relaxation}, and is
referred to as the ``SDP'' method in text and figures. 
The local method consists of using the Max-Mixture method~\cite{olsonInferenceNetworksMixtures2013}, applied to the objective function
of Problem~\eqref{prob:prob_in_terms_of_x}, with the association variables analytically eliminated~\cite[Chap.~4]{doherty2023lifelong}. 
The data associations are solved for implicitly and recovered from the continuous robot states upon convergence.
The solution considered the ``true'' solution is obtained by initializing the Max-Mixture baseline at the ground-truth robot states. 
The Gauss-Newton method is used, with a right perturbation Lie group formulation used to
handle the orientations. The same loss function is used as for the relaxation~\eqref{prob:prob_in_terms_of_x},
with the Frobenius norm handled  using the approach outlined in~\cite{dellaert2020shonan}.
The baseline algorithm, termed Max-Mix DR, consists of the Max-Mixture baseline initialized at the dead-reckoned state estimates.  
\revisionForbes{
The proposed SDP relaxation only yields a certifiably optimal result when it is tight and the obtained solution
is feasible for the non-relaxed problem, with $\rank (\mbf{Z}) = 2$. 
This is measured using the eigenvalue ratio of the second and third-largest eigenvalues of $\mbf{Z}$. 
In an ideal scenario, this eigenvalue ratio is infinite, and the matrix rank is exactly two.
In practice, a numerical cutoff threshold is used due to the numerical nature of the solver. 
An eigenvalue ratio threshold of $10^6$ is used in this work,  
following~\cite{holmes2024weighted}.
}
\revision{The experiments were run on an OptiPlex 7000 workstation with an Intel Core i9-12900K processor
and 32 GB of memory.}
\subsection{Metrics}
Position and data association errors are used to quantify algorithm performance.
The position error for a given trial is computed as the absolute trajectory error (ATE) quantified using the root-mean-square error (RMSE)~\cite{zhang2018quantitative}, 
$\text{ATE}_{\text{pos}}=\frac{1}{\NumSteps} \sum_{\TimeIdx}^\NumSteps \norm{\mbfhat{r}_i-\mbfbar{r}_i}_2$.
where
$\mbfhat{r}_i$ and $\mbfbar{r}_i$ are the estimated and true robot positions at timestep $i$. 
When computing $\text{ATE}_{\text{pos}}$ for many trials, such as for trials corresponding to given noise values,
the median across the trials is used.  
The data association error is computed as a binary quantity, corresponding to whether all the data associations are computed correctly in
the problem. When computing the data association error for many trials, the fraction of trials where the data associations
are computed correctly is used. 
\subsection{Simulated Example}
\label{sec:sim_results}
\subsubsection{Setup}
Ground truth data for simulated examples was generated by first generating $\NumSteps$ random poses on $SE(2)$
as $\mbf{T}_\TimeIdx = \exp(\mbsxi^\wedge)$, with $\mbsxi^\trans=\begin{bmatrix}
    \xi^\phi & \mbsxi^{\text{r}^\trans}
\end{bmatrix}$, with
$\xi^\phi$ generated using the uniform distribution
$\xi^\phi\sim \mc{U}(0, 2\pi)$ and
$\mbsxi^{\text{r}}$ generated from the Gaussian distribution
$\mbsxi^{\text{r}}\sim \Gaussian{\mbsxi^{\text{r}}; \mbf{0}, \eye}$.
The exponential map $\exp(\mbsxi^\wedge)$ is associated with the $SE(2)$ matrix Lie group~\cite{sola2021micro}.
Landmarks are generated on a grid with $\lndmrk_j\sim \mc{U}([0, 10]\times [0, 10])$.
Relative pose and landmark measurements are generated from the ground truth and corrupted by noise
to be used in the estimator.  
The same approach for parametrizing relative pose measurement noise is used as in~\cite{holmes2023efficient}. 
Baseline noise levels are set separately for the orientation and position components,
then scaled by a multiplier $m_{\text{noise, rel}}$. The multiplier is used as a measure of the noise. 
Relative orientation noise is parametrized by 
$1/\kappa$ of~\eqref{eq:loss_odom}. The value of $1/\kappa$ increases with higher noise. 
The relative pose position component has covariance $\stddev_r^2 \eye$. 
The base values of $1/\kappa$ and $\stddev_r^2$ are set to
$0.01 \frac{1}{\text{rad.}^2}$ and $0.745 ~\text{m}^2$. 
The parameters used for the Monte Carlo simulation runs are given by
$(\NumPosesLIW, \NumLandmarksLIW, m_{\text{noise}}, \stddev_p^2) \in (3, 5)\times (2,3)
\times (0.1, 1, 10, 20, 30, 40, 50, 60) \times (0.5, 1, 2, 3, 4, 5)$,
with 10 Monte Carlo trials generated for each
parameter configuration, for a total of 1920 trials.
The relative landmark position measurement is directly parametrized by the scalar variance multiplied
by identity covariance, 
$\stddev_p^2 \eye$. 
Priors on the first pose were set to the ground truth with $\kappa_{\text{prior, rot}}=100$ and $1/\stddev_{\text{prior, pos}}^2=0.01$.
To evaluate optimality of the proposed approach, when state errors are computed, unless explicitly stated otherwise, they are computed
\emph{with respect to a local method result initialized at the ground truth}, since the ground truth may not correspond to the global
objective minimum. 
\subsubsection{Discussion}
\begin{figure}
    \centering
    \includegraphics[width=\columnwidth]{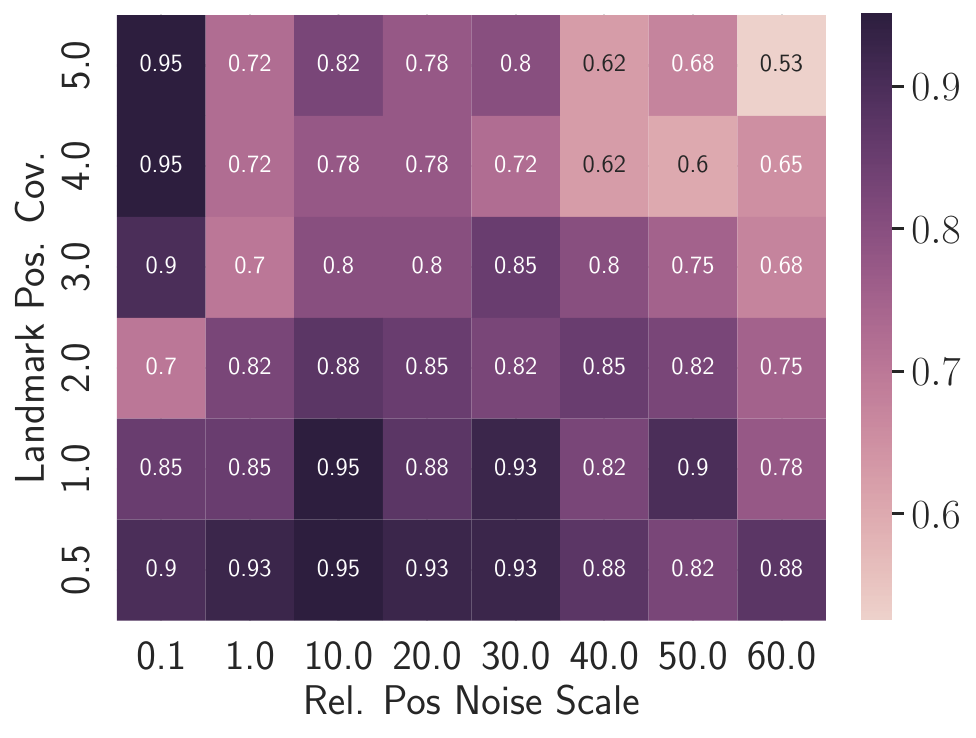}
    \caption{
        Fraction of tight cases in simulated example, evaluated using a threshold of $\lambda_2/\lambda_3\geq10^6$.
        The breakdown occurs at a relative pose measurement noise scale of around 30 and a landmark position
        noise scale of 3. 
    }
    \label{fig:simulation_tightness_fraction}
\end{figure}

\begin{figure}
    \centering
\vspace{-0.5cm}
    \includegraphics[width=\columnwidth]{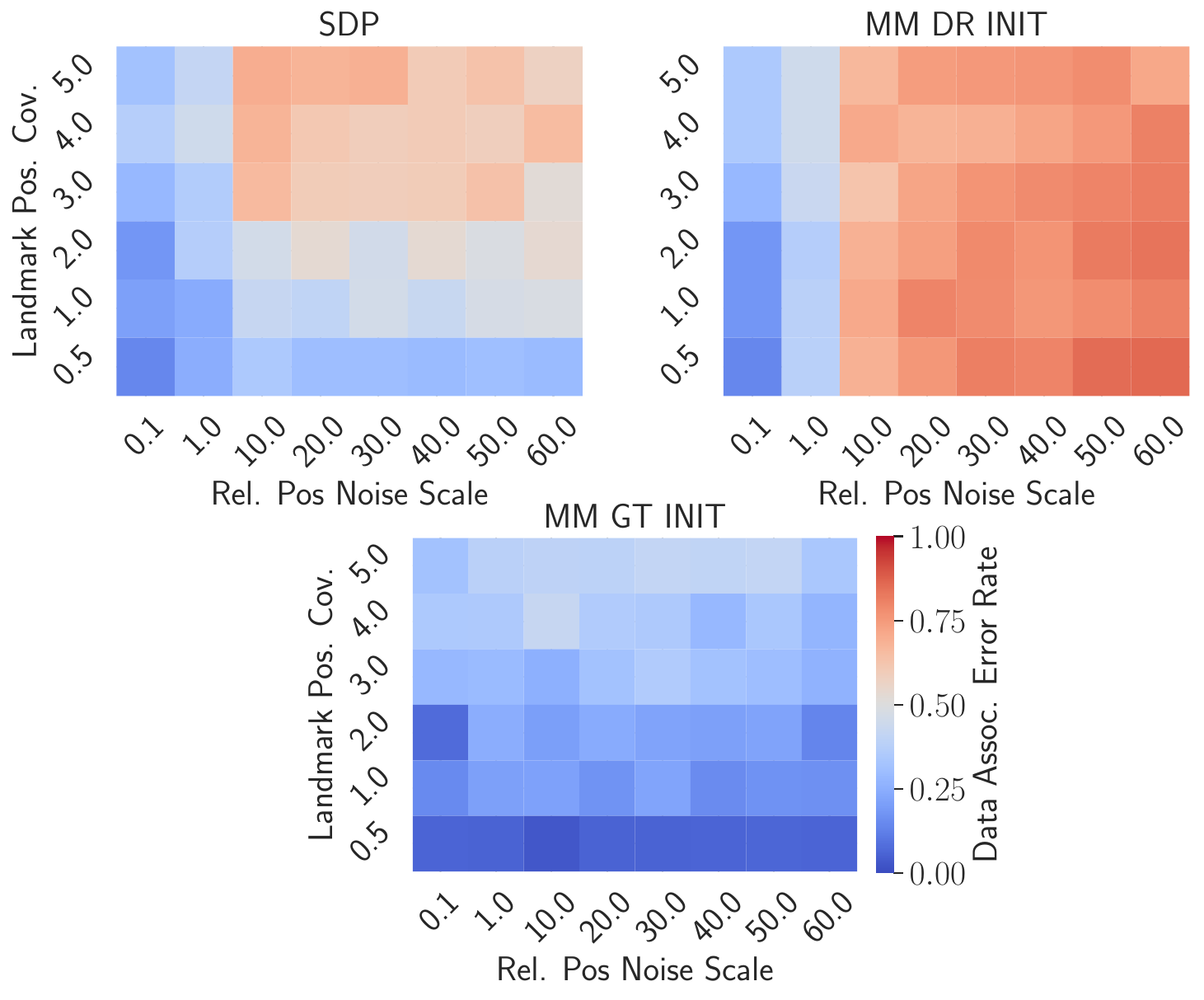}
    \caption{
        Evaluation of data association error rate for tight cases in simulated example. For each noise level combination, the corresponding
        heatmap value denotes the fraction of trials in which at least one data association is incorrect. 
    }
    \label{fig:simulation_tight_data_association}
\end{figure}

\begin{figure}
    \centering
    \includegraphics[width=\columnwidth]{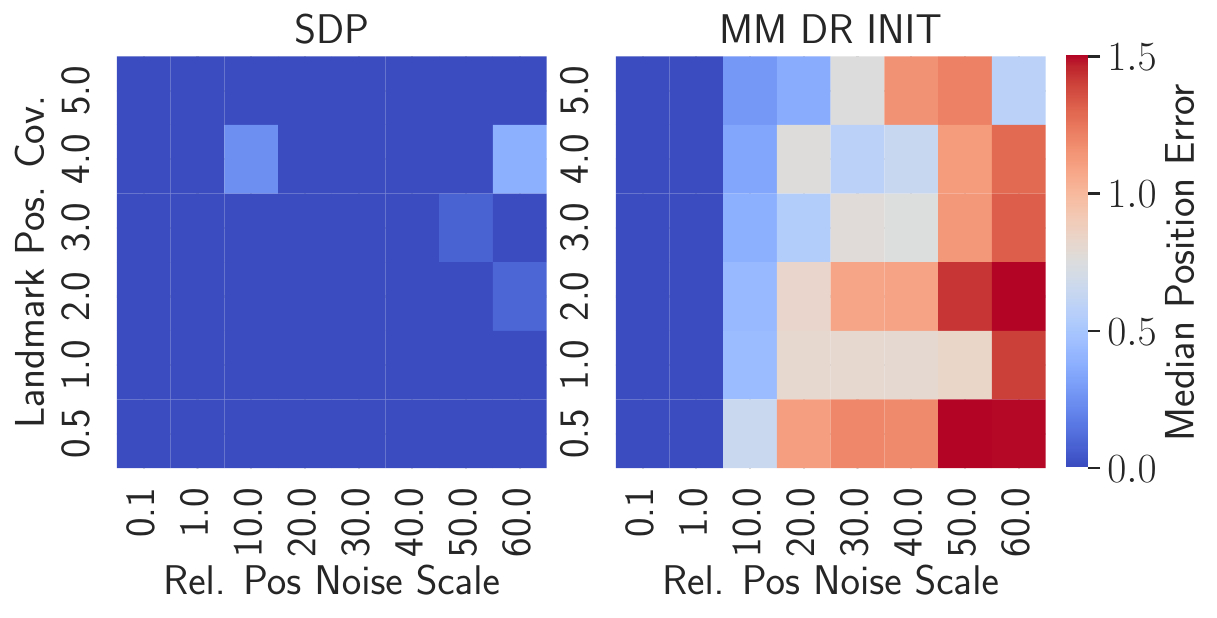}
    \caption{
        Evaluation of position error for tight cases in simulated example. Each entry in the heatmap corresponds to the median
        position error for trials for the corresponding noise values. 
    }
    \label{fig:simulation_tight_position_error}
\end{figure}

The proposed method is tight for reasonable noise levels, as demonstrated in Figure~\ref{fig:simulation_tightness_fraction}.
The method starts breaking down to around 60\% tightness rate for a landmark covariance of 4 m$^2$ and 
a relative pose noise multiplier of 40. 
\revision{
The expected behavior is for tightness fraction to go down with noise.     
The trend is not perfect, as particularly for $\text{Rel. Pos Noise Scale=1}$, the tightness fraction decreases to $0.7$
before going up again and resuming the expected trend, which is attributed to experimental spread. 
However,
}
the general trend is as expected, with tightness rates going down with increasing landmark
measurement noise as well as increasing relative pose measurement noise.
The data association correctness for tight cases is presented in Figure~\ref{fig:simulation_tight_data_association},
where the heatmaps plot the fraction of cases where at least one
computed data association did not match the original data associations versus the relative pose and relative landmark position measurement
noise levels.
For the tight cases, it is expected that the SDP method attains the correct data associations, while the
Max-Mix DR performs worse, especially with increasing noise levels. Furthermore, it is expected that
the Max-Mix GT attains the correct data associations in all cases.
While it is observed that the Max-Mix DR does perform worse than SDP,
neither SDP nor Max-Mix GT attain the ground truth data associations in all cases.  
This is because the minimum of the objective function with \emph{noisy} measurements of Problem~\eqref{prob:original_problem} does not
necessarily have the same
data associations as the ground truth \emph{noiseless} setup. 
Similarly, the SDP solution does not always match the Max-Mix GT solution because the
ground truth initialization might not attain the global minimum of the problem.
This was supported by finding that, in tight cases, the cost attained by SDP was always lower than
that of Max-Mix GT.
Furthermore, the improved data association obtained by SDP compared to Max-Mix DR 
is shown to improve position error in Figure~\ref{fig:simulation_tight_position_error}. 
\revision{
The drawback of the proposed implementation is scalability. A runtime comparison is
shown in Table~\ref{tab:solver_time}, where the proposed method has a median runtime of 32.81 seconds
for the case of 5 poses and 3 landmarks. Generic SDP solvers such as \mosek do not scale well past medium-size problems,
a threshold that is reached quickly since the problem size scales linearly with the amount of discrete variables added.
}
%
\begin{table}[ht]
\centering
\caption{Median Solver Time for Simulation Experiment (seconds)}
\begin{tabular}{|c|c|c|c|c|}
\hline
\multirow{2}{*}{Landmarks} & \multicolumn{2}{c|}{3 Poses} & \multicolumn{2}{c|}{5 Poses} \\
\cline{2-5}
& SDP & MM DR INIT & SDP & MM DR INIT \\ \hline
2 & 0.64 & $\mbf{0.03}$ & 2.19 & $\mbf{0.04}$ \\ \hline
3 & 6.14 & $\mbf{0.03}$ & 32.81 & $\mbf{0.06}$ \\ 
\hline
\end{tabular}
\label{tab:solver_time}
\end{table}
%
\subsection{Lost in the Woods Dataset}
\label{sec:lost_in_the_woods}
\subsubsection{Setup}
The Lost in the Woods dataset~\cite{lostInTheWoodsDataset} is used to validate the proposed approach on real data. It consists of a wheeled robot driving in a forest
of plastic tubes that are treated as landmarks. The robot receives wheel odometry measurements that provide forward and angular velocity measurements. 
A laser rangefinder provides range-bearing measurements to the landmarks, whose positions are provided in the dataset. 
The wheel odometry measurements are integrated to provide relative pose measurements using the body frame velocity process model~\cite{cossette2020}, while the range-bearing measurements are converted to
relative position measurements.
The orientation component of the relative pose cost~\eqref{eq:loss_odom} arises as a log-likelihood of an isotropic Langevin noise distribution on $SO(2)$~\cite{rosen2019sesync}. 
Given a set of pairs of
true relative orientations and corresponding measured relative orientations computed from odometry, the parameter $\kappa$ in~\eqref{eq:loss_odom} is computed by reverse-engineering
the isotropic Langevin distribution sampling procedure in~\cite{rosen2019sesync}. 
Priors on the first pose were set to the ground truth with $\kappa_{\text{prior, rot}}=100$ and $\stddev_{\text{prior, pos}}^2=0.01$.
The relative pose position components and the relative landmark position measurement noise covariances are also computed using the measurements compared to the ground-truth data. 
The dataset is used to extract subsequences, parametrized by the number of poses $\NumPosesLIW$, the number of landmarks used $\NumLandmarksLIW$, 
as well as the spacing between poses $\PoseSpacingLIW$, measured in seconds. The higher the spacing between the poses, the more error accumulates for the
relative pose measurements obtained by integrating the odometry. The subsequences extracted for
each dataset parameter setting $(\NumPosesLIW, \NumLandmarksLIW, \PoseSpacingLIW)$ are non-overlapping. 
While the dataset provides 17 landmarks, only a subset is used for each sequence. For a set amount of landmarks, $\NumLandmarksLIW$,
the landmarks used are chosen that are visible at the largest amount of the selected subsequence timesteps. 
The Monte Carlo trials are run for parameter configurations
given by
$(\NumPosesLIW, \NumLandmarksLIW, \PoseSpacingLIW) \in (3, 5)\times (2,3) \times (20,40,60)$. 
The overall dataset is 20 minutes long. For each parameter set $(\NumPosesLIW, \NumLandmarksLIW, \PoseSpacingLIW)$, the first subsequence
starts at $t=0$, with each following subsequence starting at the end of the previous one. 
For each combination of $\NumPosesLIW, \PoseSpacingLIW$, the number of subsequences is chosen to be the maximum allowed for by
the dataset length. 
\subsubsection{Discussion}
\begin{figure}
    \centering
    \includegraphics[width=0.8\columnwidth]{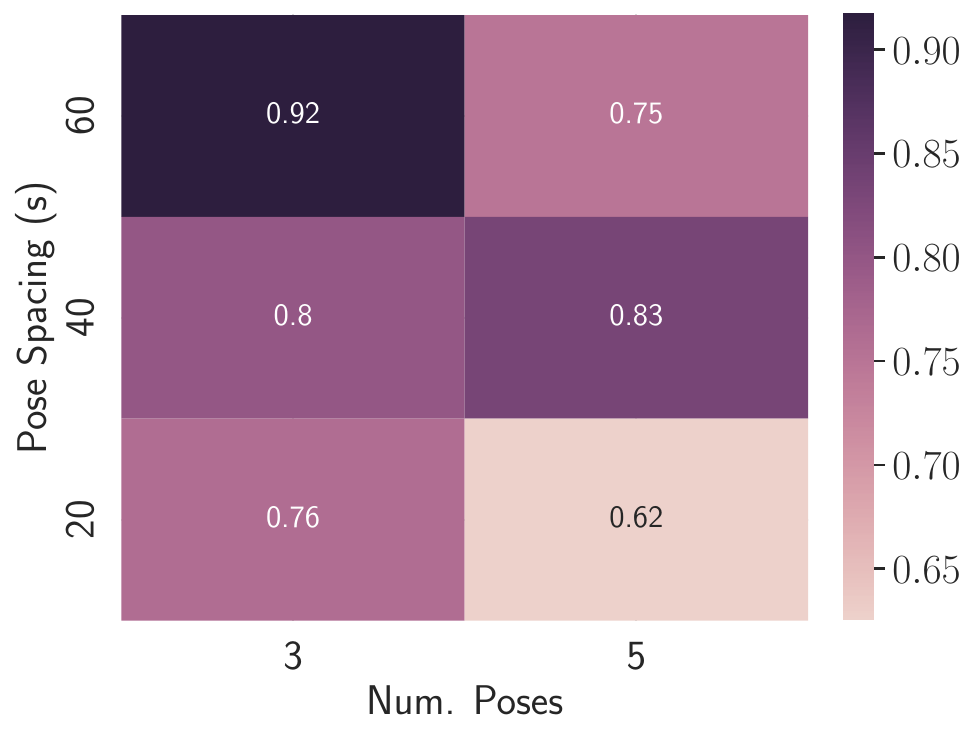}
    \caption{
        Fraction of tight cases in Lost in the Woods, evaluated using a threshold of $\lambda_2/\lambda_3\geq10^6$. 
        While the method is tight in the majority of cases, there is no clear relationship demonstrated between an increase in pose spacing,
        essentially corresponding to odometry noise, and the tightness of the relaxation.
    }
    \label{fig:liw_tightness_fraction}
\end{figure}

\begin{figure}
    \centering
    \includegraphics[width=\columnwidth]{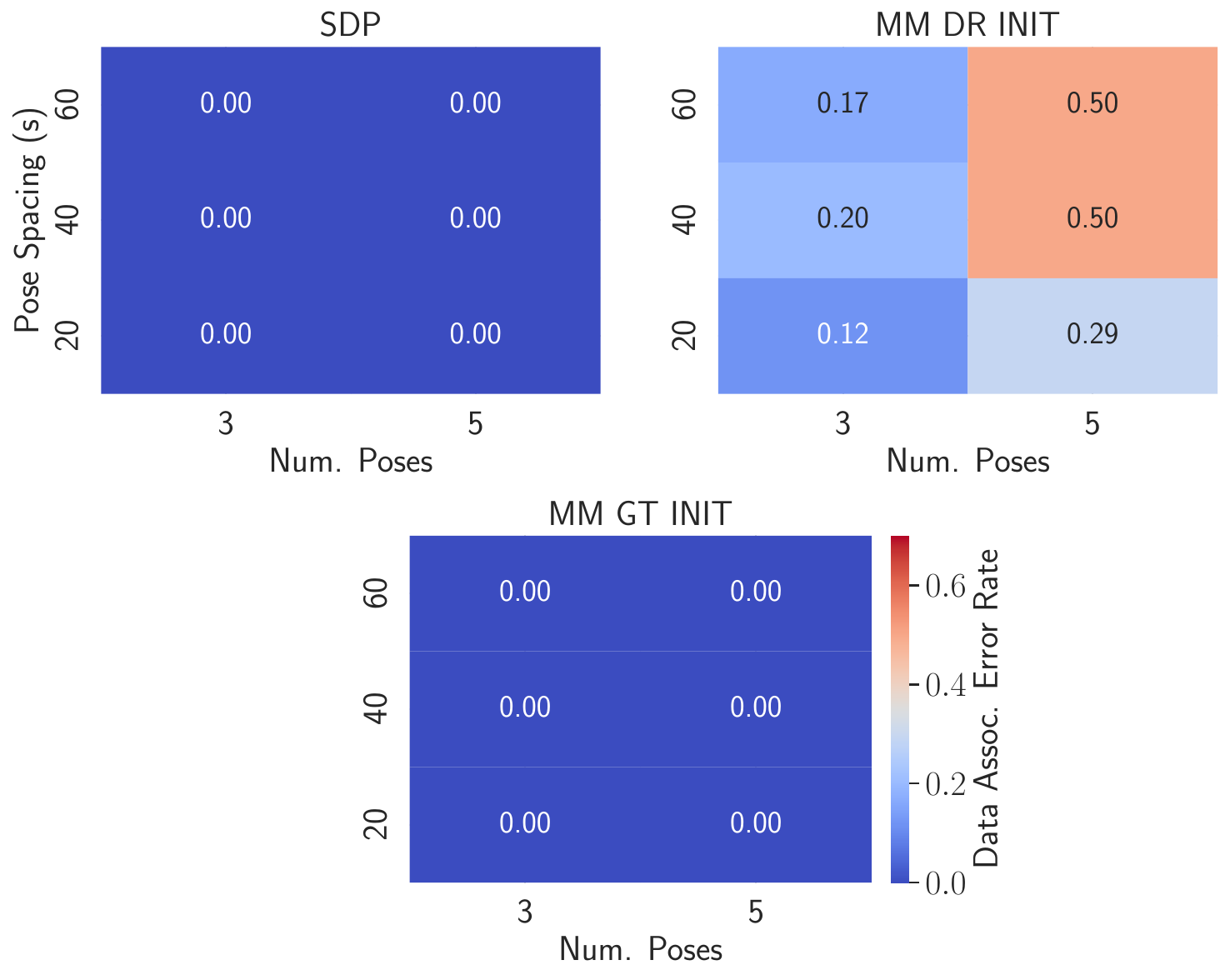}
    \caption{
        Evaluation of data association correctness for the Lost in the Woods example.
        This includes all results, both tight and nontight.  
    }
    \label{fig:liw_data_association_comparison_all}
\end{figure}

The parameters used for the $x$ and $y$ axes are different for the experiment section compared to the simulated examples.
Since the landmark measurement noise is fixed, the varied parameters consist of $\NumPosesLIW$ and $\PoseSpacingLIW$.
The $\PoseSpacingLIW$ essentially corresponds to the odometry measurement noise, while $\NumPosesLIW$
corresponds to increasing the problem size and trajectory length. 
Thus, increasing either $\PoseSpacingLIW$ or $\NumPosesLIW$ is expected to worsen the quality of the dead-reckoned initialization. 
Furthermore, increasing $\PoseSpacingLIW$ is expected to degrade the tightness of the relaxation.   
The fraction of tight cases for each parameter configuration is presented in Fig.~\ref{fig:liw_tightness_fraction}.
While the method is tight in the majority of cases, it is hard to extract a clear pattern of worsening tightness with noise in this setup. 
This can be ascribed to the fewer amount of subsequences used in this setup compared to the simulation. In Fig.~\ref{fig:simulation_tightness_fraction},
each square corresponds to 40 Monte Carlo trials, while in Fig.~\ref{fig:liw_tightness_fraction} the number of subsequences used
ranges from 21 for $\PoseSpacingLIW=20, \NumPosesLIW=3$ to only four for the $\PoseSpacingLIW=60, \NumPosesLIW=5$ case. 
The data association results are presented in
Fig.~\ref{fig:liw_data_association_comparison_all},
where both tight and nontight results are plotted together. This is unlike the simulated case where
only the tight results were plotted. In this experiment, both tight and nontight
solutions were shown to yield good results. 
In both tight and nontight cases, SDP matches Max-Mix GT in all cases,
while Max-Mix DR has a very high failure rate for $\NumPosesLIW=5$.
This is expected, since with a higher number of poses, the quality of the dead-reckoned initialization worsens leading to degradation of
performance of the local method that is dependent on the initialization.
The position error in Table~\ref{tab:liw_pos_error}
follows the same trend as the data association plot in Fig.~\ref{fig:liw_data_association_comparison_all}, with position errors increasing for $\NumPosesLIW=5$, showing that the incorrect data association of the local method
also leads to a degradation of the estimated robot state.
The position error is computed with respect to the trajectory computed by the Max-Mix GT method.
A scalar position error is computed for each trajectory, and the median for
each parameter configuration is shown in the table. 

\begin{table}[ht]
\centering
\tiny
\caption{Median Position Error For Lost in The Woods Dataset}
\begin{tabular}{|c|c|c|c|c|}
\hline
\multirow{2}{*}{Pose Spacing (s)} & \multicolumn{2}{c|}{3 Poses} & \multicolumn{2}{c|}{5 Poses} \\
\cline{2-5}
& SDP & MM DR INIT & SDP & MM DR INIT \\
\hline
20 & $1.8\!\times\!10^{-5}$ & $\mbf{2.4\!\times\!10^{-11}}$ & $5.3\!\times\!10^{-5}$ & $\mbf{3.2\!\times\!10^{-11}}$ \\
\hline
40 & $7.8\!\times\!10^{-6}$ & $\mbf{2.0\!\times\!10^{-11}}$ & $\mbf{5.5\!\times\!10^{-5}}$ & $1.2\!\times\!10^{-1}$ \\
\hline
60 & $8.9\!\times\!10^{-6}$ & $\mbf{7.5\!\times\!10^{-11}}$ & $\mbf{1.8\!\times\!10^{-5}}$ & $1.7\!\times\!10^{-1}$ \\
\hline
\end{tabular}
\label{tab:liw_pos_error}
\end{table}
\FloatBarrier

\section{Conclusion}
\label{sec:conclusion}
This paper proposes a relaxation-based globally optimal method for robot localization. 
The relaxation is tight in the majority of cases in simulation, and the majority of cases in real-world experiment.
While offering global optimality in cases where tightness is attained, 
the scalability of the algorithm is limited by the difficulty in solving large-scale SDPs.
While the scaling of semidefinite program solvers is generally still prohibitive, advances have been made for specific
problem instances, particularly for SDPs where the solution is of
low rank~\cite{rosen2019sesync, duembgen2025exploitingchordalsparsityfast, papalia2024rangeaided,yang2023outlier}, and
future SDP solvers may allow scaling of the proposed method beyond the small problem instances analyzed in this paper.
Future work also includes development of certification schemes based on the relaxation. 
\revision{The use of a null hypothesis to account for misidentified landmarks is also a promising direction of future research,
where, similarly to~\cite{poschmann2020factor}, outlier measurements are assigned to a dummy landmark. 
The null hypothesis would make use of the possibility of ``many-to-one'' assignment of measurements to landmarks. }
Furthermore, while this paper specifically considers the setting of robot localization, the overall approach
and redundant constraints used are applicable to other settings with unknown data association, investigating
which is another topic of future work. 

\printbibliography
\onecolumn
\appendix 

The appendix provides details on the construction of the cost and constraint matrices. 
In turn, the following subproblems are considered: The continuous subproblem, the discrete subproblem, 
and the combination of the two. 
\subsection{Continuous Problem}
A homogenization variable $\mbf{H}\in \{-\eye_{2\times 2}, \eye_{2\times 2}\}$, which can be enforced quadratically,
is introduced to be able to write cost and constraint terms
that are linear in $\mc{X}$ and $\mbs{\discVar}$. 
The impact of homogenization is discussed in~\cite{cifuentes2022local}. 
To fix ideas, especially related to cost functions and constraints, $\mbf{H}=\eye_{2\times 2}$
can be assumed. 
The direction cosine matrix variables may be concatenated 
to yield $\OverallRotVar=
\begin{bmatrix}
    \mbf{C}_1 & \cdots & \mbf{C}_{\TimeIdx} & \cdots & \mbf{C}_{\NumSteps} 
\end{bmatrix}$, 
while the position variables may be concatenated as 
$\OverallPosVar=\begin{bmatrix} \mbf{r}_1 & \cdots & \mbf{r}_{\TimeIdx} & \cdots & \mbf{r}_{\NumSteps}\end{bmatrix}$.
The continuous problem variable is denoted $\ContVar$ and is obtained by horizontal concatenation of
the robot states as well as of the homogenization variable
\begin{align}
    \ContVar & =
    \begin{bmatrix}
        \MatrixHomVar & \OverallRotVar & \OverallPosVar
    \end{bmatrix} \in \rnums^{2\times (2+3\NumSteps)}.
    \label{eq:cont_var_definition}
\end{align}
A cost term that depends on the continuous variable may be expressed as
$J_{\ContVar}(\ContVar)$
\begin{align}
    J_{\ContVar}(\ContVar) = \InnerProduct{\mbf{Q}_{\ContVar}}{\ContVar^\trans \ContVar}, 
\end{align}
while a constraint may be expressed as 
\begin{align}
    \InnerProduct{\mbf{A}_{\Xi}}{\ContVar^\trans \ContVar} = b,
\end{align}
where $\mbf{A}_{\Xi}$ is a matrix and $b$ a scalar. 
While there are many cost terms and constraints in the problem, the cost term and constraint index is dropped for brevity of notation. 
The matrix inner product
$\InnerProduct{\cdot}{\cdot}$ is defined 
for matrices $\mbf{A}, \mbf{B}\in \rnums^{m\times n}$
as $\InnerProduct{\mbf{A}}{\mbf{B}}=\trace(\mbf{A}\mbf{B}^\trans)=\sum_{i=1}^M \sum_{j=1}^N a_{ij}b_{ij}$. 
Importantly, the costs and constraints \emph{can only be expressed using sums of dot products of the columns of $\ContVar$} 
since, for a generic matrix $\mbf{A}$,
\begin{align}
    \InnerProduct{\mbf{A}}{\ContVar^\trans \ContVar} &= 
    \sum_{i=1}^M \sum_{j=1}^N a_{ij}(\ContVar^\trans \ContVar)_{ij} \\ 
    &=
    \sum_{i=1}^M \sum_{j=1}^N a_{ij}\ContVarCol_i^\trans \ContVarCol_j.
\end{align} 
It is worth discussing the main alternative to this approach, which consists of flattening
the variable into a single column matrix $\mbsxi \in \rnums^{1+6N}$, containing all of the robot states
and a scalar homogenization variable. 
This alternative leads to cost and constraint terms expressed using the quadratic expressions of the
form 
$\mbsxi^\trans \mbf{A} \mbsxi$. This alternative leads to a larger problem dimension and can require
a larger amount of redundant constraints on the relaxation. However, it is more flexible, since 
constraints and costs can be expressed using any scalar in the continuous variable $\ContVar$, unlike
the approach used in this work where only dot products of the columns of $\ContVar$ may be used. 
A downstream consequence of only being able to use column dot products to express the cost is
that the noise models used are less flexible. 
\subsubsection{Cost Matrix Construction}
Cost terms in the continuous variable are expressed using the quadratic form
\begin{align}
    J_{\ContVar}(\ContVar) = \InnerProduct{\mbf{Q}_{\ContVar}}{\ContVar^\trans \ContVar}.
\end{align}
For the prior and odometry terms, the entries of $\mbf{Q}_{\ContVar}$
are inserted directly into corresponding entries of the overall cost matrix. 
For the unknown data associations, they are inserted into entries of the overall cost matrix
that involve corresponding discrete variables. 
\subsubsection{Lifting a Pose to Known Landmark Measurement}
The error is the same as for the unknown landmark position measurement,
\begin{align}
    e(\mbf{C}_{ab},  \mbf{r};\lndmrk,  \mbf{y})
     & =
    \lndmrk^\trans \lndmrk + \mbf{r}^\trans \mbf{r}
    + \mbf{y}^\trans \cancel{\mbf{C}^\trans \mbf{C}} \mbf{y}
    + 2\mbf{r}^\trans \mbf{C}\mbf{y}
    - 2\mbf{r}^\trans \lndmrk
    - 2\lndmrk^\trans \mbf{C}\mbf{y}
    \\
     & =
    \lndmrk^\trans \lndmrk + \mbf{r}^\trans \mbf{r}
    + \mbf{y}^\trans \mbf{y}
    + 2\mbf{r}^\trans \mbf{C}\mbf{y}
    - 2\mbf{r}^\trans \lndmrk
    - 2\lndmrk^\trans \mbf{C}\mbf{y},
\end{align}
where subscripts are dropped for the sake of brevity.
The last term bears a bit more consideration to include it in the
matrix inner product formulation such that
\begin{align}
    \lndmrk^\trans \mbf{C}\mbf{y} & =
    \trace (\mbf{C} \mbf{y}\lndmrk^\trans)                                            \\
                                  & =
    \trace (\mbf{C} (\lndmrk \mbf{y}^\trans)^\trans)                                  \\
                                  & = \InnerProduct{\mbf{C}}{\lndmrk \mbf{y}^\trans}.
\end{align}
However, now the landmark $\lndmrk$ is known, such that the error is a function of $\mbf{C}_{ab},  \mbf{r}$ and
parametrized using $\lndmrk,  \mbf{y}$, where $\mbf{y}$ is the measured relative position of the landmark.
Using the state
$ \begin{bmatrix}
        \eye & \mbf{C} & \mbf{r}
    \end{bmatrix}$, the error may be written
\begin{align}
    e(\mbf{C}_{ab},  \mbf{r};\lndmrk,  \mbf{y})
     & =
    \lndmrk^\trans \lndmrk + \mbf{r}^\trans \mbf{r}
    + \mbf{y}^\trans \mbf{y}
    + 2\mbf{r}^\trans \mbf{C}\mbf{y}
    - 2\mbf{r}^\trans \lndmrk
    - 2\lndmrk^\trans \mbf{C}\mbf{y}               \\
     & =
    \InnerProduct{\mbf{Q}}{\mbf{X^\trans \mbf{X}}} \\
     & =
    \InnerProduct{\mbf{Q}}{
        \begin{bmatrix}
            \eye & \mbf{C} & \mbf{r}
            \\
                 & \eye    & \mbf{C}^\trans \mbf{r}
            \\
                 &         & \mbf{r}^\trans  \mbf{r}
            \\
        \end{bmatrix}
    }                                              \\
     & =
    \InnerProduct{
        \begin{bmatrix}
            \diag ([(\lndmrk^\trans \lndmrk +\mbf{y}^\trans \mbf{y})\mbf{e}_1]) & -2\lndmrk \mbf{y}^\trans & -2\lndmrk
            \\
                                                                                &                          & 2\mbf{y}
            \\
                                                                                &                          & 1
            \\
        \end{bmatrix}
    }{
        \begin{bmatrix}
            \eye & \mbf{C} & \mbf{r}
            \\
                 & \eye    & \mbf{C}^\trans \mbf{r}
            \\
                 &         & \mbf{r}^\trans  \mbf{r}
            \\
        \end{bmatrix}
    }
\end{align}

\subsubsection{Lifting a Relative Pose Measurement}
Given a relative pose measurement such that
\begin{align}
    \mbf{T}_{k+1} & = \mbf{T}_k \Delta \mbf{T} \\
    \begin{bmatrix}
        \mbf{C}_{k+1} & \mbf{r}_{k+1} \\
        \mbf{0}       & 1
    \end{bmatrix}
                  & =
    \begin{bmatrix}
        \mbf{C}_{k} \Delta \mbf{C} & \mbf{r}_k+\mbf{C}_{k} \Delta \mbf{r} \\
        \mbf{0}                    & 1
    \end{bmatrix},
\end{align}
the first error corresponds to the translation error and is given by
\begin{align}
    J_r & = \norm{\mbf{r}_{k+1} - \mbf{r}_k-\mbf{C}_{k} \Delta \mbf{r}}_2^2                                                     \\
        & =(\mbf{r}_{k+1} - \mbf{r}_k-\mbf{C}_{k} \Delta \mbf{r})^\trans (\mbf{r}_{k+1} - \mbf{r}_k-\mbf{C}_{k} \Delta \mbf{r}) \\
        & =\mbf{r}_{k+1}^\trans \mbf{r}_{k+1} + \mbf{r}_k^\trans \mbf{r}_k + \Delta \mbf{r}^\trans \Delta \mbf{r}
    +2 \mbf{r}_k^\trans \mbf{C}_{k} \Delta \mbf{r}
    -2\mbf{r}_{k+1}^\trans\mbf{C}_k \Delta \mbf{r}
    -2\mbf{r}_{k+1}\mbf{r}_k.
\end{align}
For $\mbf{X}=\begin{bmatrix}
        \mbf{C}_k & \mbf{r}_k & \mbf{r}_{k+1}
    \end{bmatrix}$, this yields the following cost,
\begin{align}
    J_r & =
    \InnerProduct{\mbf{Q}}{\mbf{X}^\trans \mbf{X}} \\
        & =
    \InnerProduct{\mbf{Q}}{\begin{bmatrix}
            \mbf{C}_k^\trans \\ \mbf{r}_k^\trans \\ \mbf{r}_{k+1}
        \end{bmatrix}
        \begin{bmatrix}
            \mbf{C}_k & \mbf{r}_k & \mbf{r}_{k+1}
        \end{bmatrix}}                \\
        & =
    \InnerProduct{\mbf{Q}}{
        \begin{bmatrix}
            \mbf{C}_k^\trans  \mbf{C}_k    & \mbf{C}_k^\trans  \mbf{r}_k    & \mbf{C}_k^\trans \mbf{r}_{k+1}
            \\
            \mbf{r}_k^\trans  \mbf{C}_k    & \mbf{r}_k^\trans \mbf{r}_k     & \mbf{r}_k^\trans \mbf{r}_{k+1}     \\
            \mbf{r}_{k+1}^\trans \mbf{C}_k & \mbf{r}_{k+1}^\trans \mbf{r}_k & \mbf{r}_{k+1}^\trans \mbf{r}_{k+1}
        \end{bmatrix}
    }                                              \\
        & =
    \InnerProduct{
        \begin{bmatrix}
            \mbf{0} & 2\Delta \mbf{r} & -2\Delta \mbf{r}
            \\
                    & 1               & -2               \\
                    &                 & 1
        \end{bmatrix}
    }{
        \begin{bmatrix}
            \mbf{C}_k                      & \mbf{r}_k                      & \mbf{r}_{k+1}
            \\
            \mbf{r}_k^\trans  \mbf{C}_k    & \mbf{r}_k^\trans \mbf{r}_k     & \mbf{r}_k^\trans \mbf{r}_{k+1}     \\
            \mbf{r}_{k+1}^\trans \mbf{C}_k & \mbf{r}_{k+1}^\trans \mbf{r}_k & \mbf{r}_{k+1}^\trans \mbf{r}_{k+1}
        \end{bmatrix}
    }.
\end{align}
The second error corresponds to the rotation error and is given by
\begin{align}
    J_c & = \norm{\mbf{C}_{k+1}-\mbf{C}_k \Delta \mbf{C}}_F^2                                                                                     \\
        & = \trace \left(\left(\mbf{C}_{k+1}-\mbf{C}_k \Delta \mbf{C} \right) \left(\mbf{C}_{k+1}-\mbf{C}_k \Delta \mbf{C} \right)^\trans \right) \\
        & =
    2\trace (\eye) -2\trace (\mbf{C}_{k+1}^\trans \mbf{C}_k \Delta \mbf{C})                                                                       \\
        & =
    2\trace(\eye - \Delta \mbf{C}\mbf{C}_{k+1}^\trans \mbf{C}_k)                                                                                  \\                                                                      \\
\end{align}
Making use of $\trace (\mbf{A}\mbf{B}^\trans)=\sum_i \sum_j \mbf{A}_{ij}\mbf{B}_{ij}=\InnerProduct{\mbf{A}}{\mbf{B}}$
allows to write
\begin{align}
    J_c & =
    2\trace(\eye - \Delta \mbf{C}\mbf{C}_{k+1}^\trans \mbf{C}_k)             \\
        & = 2\left(
    \trace \eye
    - \InnerProduct{\Delta \mbf{C}}{ \mbf{C}_k^\trans \mbf{C}_{k+1} }\right) \\
        & =
    \InnerProduct{
        \mbf{Q}
    }{
        \begin{bmatrix}
            \eye             \\
            \mbf{C}_k^\trans \\
            \mbf{C}_{k+1}^\trans
        \end{bmatrix}
        \begin{bmatrix}
            \eye      &
            \mbf{C}_k &
            \mbf{C}_{k+1}
        \end{bmatrix}
    }                                                                        \\
        & =
    \InnerProduct{
        \begin{bmatrix}
            2 \eye &         &                   \\
                   & \mbf{0} & -2 \Delta \mbf{C} \\
                   &         & \mbf{0}
        \end{bmatrix}
    }{
        \begin{bmatrix}
            \eye                 & \mbf{C}_k                      & \mbf{C}_{k+1}                      \\
            \mbf{C}_k^\trans     & \mbf{C}_k^\trans \mbf{C}_k     & \mbf{C}_k^\trans \mbf{C}_{k+1}     \\
            \mbf{C}_{k+1}^\trans & \mbf{C}_{k+1}^\trans \mbf{C}_k & \mbf{C}_{k+1}^\trans \mbf{C}_{k+1}
        \end{bmatrix}
    }.
\end{align}
%
\subsubsection{Lifting a Prior}
Given a prior such that the error is given by
\begin{align}
    J_p & = \norm{\mbf{x}-\mbfcheck{x}}_2^2              \\
        & =
    (\mbf{x}-\mbfcheck{x})^\trans (\mbf{x}-\mbfcheck{x}) \\
        & =
    \mbf{x}^\trans \mbf{x} - 2\mbfcheck{x}^\trans \mbf{x} +
    \mbfcheck{x}^\trans\mbfcheck{x}
\end{align}
for $\mbf{X}=\begin{bmatrix}
        \eye & \mbf{x}
    \end{bmatrix}$, this yields the following cost
\begin{align}
    J_p & =
    \mbf{x}^\trans \mbf{x} - 2\mbfcheck{x}^\trans \mbf{x} +
    \mbfcheck{x}^\trans\mbfcheck{x} \\
        & =
    \InnerProduct{\mbf{Q}}{\mbf{X}^\trans \mbf{X}}
    \\
        & =
    \InnerProduct{
        \begin{bmatrix}
            \mbfcheck{x}^\trans\mbfcheck{x} & - 2\mbfcheck{x}^\trans \\
                                            & 1
        \end{bmatrix}
    }{
        \begin{bmatrix}
            \eye    & \mbf{x}                \\
            \mbf{x} & \mbf{x}^\trans \mbf{x}
        \end{bmatrix}
    }.
\end{align}
\subsubsection{Constraint Matrix Definition}
The continuous variable constraints enforce orthonormality of the direction cosine matrix (DCMs)
in the relaxation. 
For brevity, the constraints are given for the specific subvariables considered. 
In the overall problem, they are then inserted directly into the entries of the overall
constraint matrix corresponding the subvariables.  
Symmetric constraint matrices are used in SDPs. Any non-symmetric constraint matrix
may be symmetrized as $\mbf{A}_{\text{sym}}=(\mbf{A}+\mbf{A}^\trans)/2$.
Given a DCM $\mbf{C}$, the continuous variable constraints are of the form
\begin{align}
    \InnerProduct{\mbf{A}}{
    \begin{bmatrix}
    \MatrixHomVar & \mbf{C} \\   
    \mbf{C}^\trans & \mbf{C}^\trans \mbf{C}
    \end{bmatrix}
    } = 0,
\end{align}
where the homogenization variable $\MatrixHomVar$ can be thought of as identity. 
The matrix $\mbf{A}$ is used as a generic constraint matrix in this section. 
The continuous constraint matrices then correspond to 
\begin{enumerate}
    \item Orthonormality. For each column index combination  $i, j \in (1,2)\times (1,2)$,
    a constraint matrix is added with the following structure. 
    \begin{itemize}
        \item If $i\neq j$, 
        constraint matrix $\mbf{A}$ has 
        $\sliceMat{\mbf{A}}{\mbf{c}_1}{\mbf{c}_2}=1$ corresponding to orthogonality
        of different columns of $\mbf{C}$. 
        \item If $i=j$,
        constraint matrix $\mbf{A}$ has 
        $\sliceMat{\mbf{A}}{\mbf{c}_1}{\mbf{c}_2}=1$, and 
        $\sliceMat{\mbf{A}}{\mbf{h}_1}{\mbf{h}_1}=-1$, corresponding to unit norm
        for each column of $\mbf{C}$.
    \end{itemize}
    \item Structure of two-dimensional DCM. For the planar case, the DCM has the structure
    $\mbf{C}=
    \begin{bmatrix}
        c & -s \\ 
        s & c
    \end{bmatrix}
    $. Two additional constraints may then be added, 
    \begin{itemize}
        \item Constraint matrix $\mbf{A}$ with $\sliceMat{\mbf{A}}{\mbf{c}_1}{\mbf{h}_1}=1$, 
        $\sliceMat{\mbf{A}}{\mbf{c}_2}{\mbf{h}_2}=-1$,
        \item Constraint matrix $\mbf{A}$ with $\sliceMat{\mbf{A}}{\mbf{c}_1}{\mbf{h}_2}=1$, 
        $\sliceMat{\mbf{A}}{\mbf{c}_2}{\mbf{h}_1}=1$. 
    \end{itemize}
\end{enumerate} 
\subsection{Discrete Problem}
\label{appendix:sec:discrete}
The discrete variables arise from the unknown data association cost given by
\begin{align}
    J_{\text{uda, lndmrk}}(\mc{X}) &= 
    \sum_{\TimeIdx=1}^{\NumSteps}
    \sum_{\LandmarkMeasIdx_{\TimeIdx}=1}^{\NumLandmarkMeas_{\TimeIdx}}
    \sum_{\LandmarkIdx=1}^{\NumLandmarks}
    \discVar_{\TimeIdx \LandmarkMeasIdx \LandmarkIdx}
    r(\mbf{C}_\TimeIdx, \mbf{r}_\TimeIdx; \lndmrk_\LandmarkIdx, \LndmrkMeas_{\LandmarkMeasIdx_{\TimeIdx}}). 
\end{align}
The outer loop is over the time indices $\TimeIdx$.
The middle loop is over the $\LandmarkMeasIdx$'th received measurement at timestep $\TimeIdx$, $\LndmrkMeas_{\LandmarkMeasIdx_{\TimeIdx}}$. 
The inner loop is over the possible data associations. 
The discrete variables are constrained to be boolean, $\theta_{\TimeIdx \LandmarkMeasIdx_{\TimeIdx} \LandmarkIdx}\in \{0, 1\}$. 
Furthermore, 
they are constrained by 
\begin{align}
    \sum_{\LandmarkIdx=1}^{\NumLandmarks}
    \discVar_{\TimeIdx \LandmarkMeasIdx \LandmarkIdx} =1, \quad \forall {\TimeIdx, \LandmarkMeasIdx}. 
\end{align}
meaning that every measurement comes from a single landmark.
The overall problem variable involves both discrete and continuous variables, and requires redundant constraints. 
These constraints in turn require quadratic constraints on the discrete variables of the form
\begin{align}
    \begin{bmatrix}
    1 & \mbs{\discVar}^\trans
\end{bmatrix} \mbf{A}_\theta 
    \begin{bmatrix}
    1 \\ \mbs{\discVar}
\end{bmatrix} =0. 
\end{align}
Constraint matrices $\mbf{A}_\theta $ may be constructed from the following observations.
Since they are used to construct \emph{redundant} constraints, it is insufficient to only use the boolean and sum constraints, and a few more are required. 
\begin{itemize}
    \item $\sum_{\LandmarkIdx=1}^{\NumLandmarks} \discVar_{\TimeIdx, \LandmarkMeasIdx, \LandmarkIdx} =1$. 
    This corresponds to $\mbf{A}_\theta$ with $\sliceMat{\mbf{A}}{1}{\discVar_{\TimeIdx, \LandmarkMeasIdx, \LandmarkIdx}}=1$ for every $j$, as well as 
    $\sliceMat{\mbf{A}}{1}{1}=-1$.
    \item $\discVar_{\TimeIdx, \LandmarkMeasIdx, \LandmarkIdx}^2-\discVar_{\TimeIdx, \LandmarkMeasIdx, \LandmarkIdx}=0$. 
    This corresponds to $\mbf{A}_\theta$ with
    $\sliceMat{\mbf{A}}{\discVar_{\TimeIdx, \LandmarkMeasIdx, \LandmarkIdx}}{\discVar_{\TimeIdx, \LandmarkMeasIdx, \LandmarkIdx}}=1$, 
    $\sliceMat{\mbf{A}}{1}{\discVar_{\TimeIdx, \LandmarkMeasIdx, \LandmarkIdx}}=-1$.
    \item $\discVar_{\TimeIdx, \LandmarkMeasIdx, \LandmarkIdx_1} \discVar_{\TimeIdx, \LandmarkMeasIdx, \LandmarkIdx_2}=0$, 
    for $\LandmarkIdx_1\neq \LandmarkIdx_2$. This corresponds to
    $\mbf{A}_\theta$ with
    $\sliceMat{\mbf{A}}{\discVar_{\TimeIdx, \LandmarkMeasIdx, \LandmarkIdx_1}}{\discVar_{\TimeIdx, \LandmarkMeasIdx, \LandmarkIdx_2}}=1$. 
    \item 
    $\discVar_{\TimeIdx, \LandmarkMeasIdx_2, \LandmarkIdx_2} \sum_{\LandmarkIdx=1}^{\NumLandmarks} 
    \discVar_{\TimeIdx, \LandmarkMeasIdx_1, \LandmarkIdx} - \discVar_{\TimeIdx, \LandmarkMeasIdx_2, \LandmarkIdx_2}=0$, 
    for $\LandmarkMeasIdx_1 \neq \LandmarkMeasIdx_2$, one constraint for each applicable $\LandmarkIdx_2$. 
    This is a premultiplication of the sum constraint for one unknown data association measurement with discrete variables
    corresponding to other unknown data association measurements, from the same timestep.
    This corresponds to
    $\mbf{A}_\theta$ constraints, one for each $\LandmarkMeasIdx_1 \neq \LandmarkMeasIdx_2$ and applicable $\LandmarkIdx_2$.
    For given $\LandmarkMeasIdx_1, \LandmarkMeasIdx_2, \LandmarkIdx_2$, the constraint matrix $\mbf{A}_\theta$ has
    $\sliceMat{\mbf{A}}{\discVar_{\TimeIdx, \LandmarkMeasIdx_2, \LandmarkIdx_2}}{\discVar_{\TimeIdx, \LandmarkMeasIdx_1, \LandmarkIdx}}=1$
    and
    $\sliceMat{\mbf{A}}{\discVar_{\TimeIdx, \LandmarkMeasIdx_2, \LandmarkIdx_2}}{1}=-1$.
\end{itemize}
\subsection{Combination of Discrete and Continuous Problems}
The overall problem contains both discrete and continuous variables with corresponding costs and constraints.
The optimization variable of interest is given by
\begin{align}
    \mbf{X} & =
    \begin{bmatrix}
        \MatrixHomVar & \mbs{\theta}^\trans \otimes \eye^{2\times 2} & \mbs{\theta}^\trans \otimes \mbs{\Xi} & \mbs{\Xi}
    \end{bmatrix}.
\end{align}
The localization problem is expressed as
\begin{mini*}|l|
    {\mbf{X}}{
        \sum_{i=1}^{\NumDiscVar}
    \InnerProduct{\mbf{X}^\trans \mbf{X}}{\mbf{Q}}
    }{}{}\tag{\text{Reform. Prob}}
    \addConstraint{\mbf{X}\in \{-\eye, \eye\} \times  \{\mbf{0}, \eye\}^{\NumDiscVar} \times }
    \addConstraint{}{\quad \left(SO(2)^\NumSteps \times (\rnums^2)^\NumSteps\right)^{(1+\NumDiscVar)}.}
\end{mini*}
The cost matrix $\mbf{Q}$ is the overall cost matrix for the problem.
The loss may be split into two parts, one solely depending on the continuous robot states $\mc{X}$,
and one depending on both $\mc{X}$ and the discrete variables $\mbs{\discVar}$. 
\begin{align}
    J(\mc{X}, \mbs{\discVar})&=J_1(\mc{X}) + J_2(\mc{X}, \mbs{\discVar}).
\end{align}
The part of the loss $J_1(\mc{X})$ depending solely on $\mc{X}$ may be written as
\begin{align}
    J_1(\mc{X}) &= 
    \sum_{i=1}^{n_{\text{f, cont}}}
    \InnerProduct{\mbf{Q}_{\ContVar, i}}{\ContVar^\trans \ContVar},
\end{align}
where $n_{\text{f, cont}}$ is the number of error terms that depend solely on the continuous variable.
The continuous variable cost matrices $\mbf{Q}_{\ContVar, i}$ are inserted directly into
the corresponding entries of the overall cost matrix $\mbf{Q}$.  
The discrete-variable dependent loss $J_2(\mc{X}, \mbs{\discVar})$
takes the form
\begin{align}
    J_2(\mc{X}, \mbs{\discVar}) = 
    \sum_{\discVarIdx=1}^{\NumDiscVar} 
    \discVar_\discVarIdx
    \InnerProduct{\mbf{Q}_{\ContVar, \discVarIdx}}{\ContVar^\trans \ContVar},
\end{align}
where $i$ now loops over the discrete variables, each of which has an associated relative landmark measurement
error expressed using $\InnerProduct{\mbf{Q}_{\ContVar, i}}{\ContVar^\trans \ContVar}$.
Each summand may be written as
\begin{align}
    \discVar_\discVarIdx
    \InnerProduct{\mbf{Q}_{\ContVar, \discVarIdx}}{\ContVar^\trans \ContVar}
    &=
    \InnerProduct{\mbf{Q}_{\ContVar, \discVarIdx}}{\ContVar^\trans \left(\discVar_\discVarIdx\ContVar\right)}.
\end{align}
Therefore, 
for each pair of columns $\ContVarCol_j, \ContVarCol_k$ of $\ContVar$, 
the corresponding parts of the overall cost matrix $\mbf{Q}$
may be set as $\sliceMat{\mbf{Q}}{\ContVarCol_j}{\theta_\discVarIdx \ContVarCol_k}
=\sliceMat{\mbf{Q}_{\ContVar, \discVarIdx}}{\ContVarCol_j}{\ContVarCol_k}$.
The discrete variable constraints of Sec.~\ref{appendix:sec:discrete} may be imposed on the overall optimization variable $\mbf{X}$
by creating a constraint matrix $\mbf{A}$ with $\sliceMat{\mbf{A}}{\mbsTheta_{i, 1}}{\mbsTheta_{j,1}}=
\sliceMat{\mbf{A}_\mbsTheta}{\discVar_i}{\discVar_j}
$. The column $\mbsTheta_{i, 1}$ corresponds to the first column of the $\theta_i \eye$ 
matrix present in the $\mbs{\theta}^\trans \otimes \eye^{2\times 2}$ section of
the optimization variable~\eqref{eq:app:column_struct_op_var}.
Similarly, the column $\mbsTheta_{j, 1}$ corresponds to the first column of the $\theta_j \eye$ 
matrix present in the $\mbs{\theta}^\trans \otimes \eye^{2\times 2}$ section of
the optimization variable~\eqref{eq:app:column_struct_op_var}. 
Furthermore, redundant constraints are created by combining the discrete and continuous variable constraints,
the details for which are present in the paper itself. 
\subsection{Redundant Constraints Corresponding to Problem Structure} 
\subsubsection{Moment Constraints}
Moment constraints are added that reflect the particular matrix structure. 
For the moment constraint presentation only, $\discVar_{\discVarIdx, 1}$ and $\discVar_{\discVarIdx, 2}$
are used to refer to the first and second columns of the matrix subblock of $\mbf{X}$
that corresponds to $\discVar_{\discVarIdx}$.
\begin{itemize}
    \item Moment constraint form 1 
    \begin{itemize}
    \item $\slicedMat{\mbf{A}}{\MatrixHomVar_i}{\discVar_\discVarIdx \ContVarCol_\ContVarIdx}{\mbf{X}}=1$, 
    \item $\slicedMat{\mbf{A}}{\discVar_\discVarIdx}{\ContVarCol_\ContVarIdx}{\mbf{X}}=-1$.
\end{itemize}
    \item Moment constraint form 2
    \begin{itemize}
        \item  $\slicedMat{\mbf{A}}{\discVar_{\discVarIdx, 1} \ContVarCol_\ContVarIdx}{\discVar_{\discVarIdx, 2}}{\mbf{X}}=1$,
        \item $\slicedMat{\mbf{A}}{\discVar_{\discVarIdx, 2} \ContVarCol_\ContVarIdx}{\discVar_{\discVarIdx, 1}}{\mbf{X}}=-1$.
    \end{itemize}
    \item Moment constraint form 3 
    \begin{itemize}
        \item $\slicedMat{\mbf{A}}{\discVar_{\discVarIdx, 1} \ContVarCol_{\ContVarIdx,1}}{\discVar_{\discVarIdx, 2} \ContVarCol_{\ContVarIdx,2}}{\mbf{X}}=1$,
        \item $\slicedMat{\mbf{A}}{\discVar_{\discVarIdx, 1} \ContVarCol_{\ContVarIdx,2}}{\discVar_{\discVarIdx, 2} \ContVarCol_{\ContVarIdx,1}}{\mbf{X}}=-1$.
    \end{itemize}
\end{itemize}
\subsubsection{Column Structure of Optimization Variable}
Due to the matrix structure of the optimization variable,
\begin{align}
    \mbf{X} & =
    \begin{bmatrix}
        \MatrixHomVar & \mbs{\theta}^\trans \otimes \eye^{2\times 2} & \mbs{\theta}^\trans \otimes \mbs{\Xi} & \mbs{\Xi}
    \end{bmatrix}, 
    \label{eq:app:column_struct_op_var}
\end{align}
where diagonal matrices
are used to encode the discrete variables, extra redundant constraints need to be imposed on the relaxation. 
Let $\discVar_{i, d_1}$ 
denote the matrix column name corresponding to the $d_1$'th column of the diagonal matrix corresponding to $\discVar_{i}$. 
Similarly, let $\discVar_{j, d_2}$ 
denote the matrix column name corresponding to the $d_2$'th column of the diagonal matrix corresponding to
the discrete variable $\discVar_{j}$, with $d_1\neq d_2$. 
Then, a redundant constraint is given by $\slicedMat{\mbf{Z}}{\discVar_{i, d_1}}{\discVar_{j, d_2}}{\mbf{X}}=0$.
This redundant constraint may be
encoded using a quadratic constraint as
$\slicedMat{\mbf{A}}{\discVar_{i, d_1}}{\discVar_{j, d_2}}{\mbf{X}}=1$.
Furthermore, additional constraints are used to impose the diagonal structure of
the blocks of the continuous variable~\eqref{eq:optimization_variable_definition} that are diagonal, 
in the same fashion as in~\cite{lajoie2019perceptual}.

\addtolength{\textheight}{-12cm}   

\end{document}